\documentclass[preprint,12pt]{elsarticle}

\usepackage{amsmath}
\usepackage{amssymb}
\usepackage{amsthm}
\usepackage{boldtensors}
\usepackage{geometry}
\usepackage{graphicx}
\usepackage{lineno}
\usepackage{microtype}
\usepackage{multirow}
\usepackage{url} 
\usepackage{xcolor}
\usepackage{tabularray}

\usepackage{subfigure}
\usepackage{algorithm}
\usepackage[noend]{algpseudocode}
\usepackage{color}
\usepackage{tabularray}
\usepackage{longtable}
 \usepackage{siunitx}
\usepackage[normalem]{ulem}

\journal{Journal of Computational and Applied Mathematics}

\begin{document}

\begin{frontmatter}



\title{Random Forest Regression Feature Importance for Climate Impact Pathway Detection}


\author[1]{Meredith G. L. Brown}
\author[1]{Matt Peterson}
\author[2]{Irina Tezaur\corref{cor1}}
\ead{ikalash@sandia.gov}
\cortext[cor1]{Corresponding author}
\author[1]{Kara Peterson}
\author[1]{Diana Bull}

\affiliation[1]{organization={Sandia National Laboratories},
            addressline={}, 
            city={Albuquerque},
            postcode={}, 
            state={NM},
            country={USA}}
\affiliation[2]{organization={Sandia National Laboratories},
            addressline={}, 
            city={Livermore},
            postcode={}, 
            state={CA},
            country={USA}}

\begin{abstract}
Disturbances to the climate system, both natural and anthropogenic, have far reaching impacts that are not always easy to identify or quantify using traditional climate science analyses or causal modeling techniques. 
In this paper, we develop
a novel technique for discovering and ranking the chain of spatio-temporal downstream impacts of a climate source, referred to herein as a source-impact pathway, using Random Forest Regression (RFR)
and SHapley Additive exPlanation (SHAP) feature importances.  Rather than utilizing RFR for classification or regression tasks (the most common use case for RFR), we propose a fundamentally new workflow in which we: (i) train random forest (RF) regressors on a set of spatio-temporal features of interest, (ii) calculate their pair-wise feature importances using the SHAP weights associated with those features, and (iii) translate these feature importances into a weighted pathway network (i.e., a weighted directed graph), which can be used to trace out and rank interdependencies between climate features and/or modalities.
Importantly, while herein we employ RFR and SHAP feature importance in steps (i) and (ii) of our algorithm, our novel workflow is in no way tied to these approaches, which could be replaced  
with \textit{any} regression method and sensitivity method.
We adopt a tiered verification approach to verify our new pathway identification methodology.  In this approach, we apply our method to ensembles of data generated by running two increasingly complex benchmarks: (i) a set of synthetic coupled equations, 
and (ii) a fully coupled simulation of the 1991 eruption of Mount Pinatubo in the Philippines performed using a modified version 2 of the U.S. Department of Energy's Energy Exascale Earth System Model (E3SMv2).  
We find that our RFR feature importance-based approach can accurately detect known pathways of impact for both test cases. 
\end{abstract}


\begin{highlights}
\item We develop a novel method that can discover and rank climate source-impact pathways.
\item The approach uniquely combines random forest regression (RFR) and feature importance.
\item We verify our method on a synthetic problem and the 1991 Mount Pinatubo eruption.
\item Our approach can accurately detect known pathways of impact for both test cases.
\end{highlights}


\end{frontmatter}



\section{Introduction}\label{sec:intro}

The Earth's climate is ever changing, with complex interactions between the atmosphere, solid Earth, hydrosphere, cryosphere and biosphere. Since disturbances, both natural and anthropogenic, in any of these spheres have far ranging impacts on life on Earth, it is critical to 
understand what we will refer to herein as ``source-impact pathways": the relationships and interactions of a set of climate variables in space-time due to an external climate forcing.  
Methods designed to identify and quantify multiple steps in a pathway by flagging which disturbances arise could greatly enhance our ability to understand high consequence ramifications with legal, political, and national security implications \cite{Burger2020}. 

The most common method for assessing impacts in the climate system is fingerprinting.  
In this approach, spatial and/or temporal patterns are established under various disturbances (i.e., greenhouse gases, aerosol loading, etc.) and matched to observations \cite{Hasselmann1993, Santer1993}.  
Although the past few years have seen extensions of fingerprinting to regional analyses \cite{bonfils2008,stott2010}, multiple variables \cite{bonfils2020, Marvel2020} and challenging problems with very small signal-to-noise ratios  \cite{Allen2003, Ribes2013, Wills2020,Weylandt:2024}, the method is designed to work within a single step.  There has been some recent work to develop conditional multi-step fingerprinting methods \cite{Wentland2024}, but this field is still in its infancy. 


While causal modeling is able to identify relationships between multiple variables 
\cite{Runge2019, Nowack:2020}, these techniques are valid for spatially-stationary signals.  Moreover, 
the high-dimensionality of climate data often limits the scalability of these methods \cite{Runge2019} (although there are recent developments towards spatially resolved causal methods (CaStLe) \cite{Nichol2024}).  
Since realistic source-impact pathways between climate variables dynamically evolve, 
off-the-shelf causal modeling methods cannot be applied to identify these relationships.


Recent years have also seen the emergence of deep learning-based methods for climate attribution, detection and impact analysis \cite{bone2022detection,mamalakis2020explainable,buckland2019using,Hart:2023,hart2024}.  These approaches typically train a neural network (NN) on an ensemble of climate data and use the NN to make predictions or as a surrogate in an inverse attribution workflow.  The primary downside of NNs is that they  
typically require massive amounts of data, and can be very costly to train, both in terms of developer time and computational requirements.  
Importantly, existing approaches do not have the capability to \textit{discover} source-impact pathways from data; instead, they require an analyst to postulate a set of possible relationships, which are then confirmed/denied.  
Additionally, existing methods typically do not provide
information about relative pathway strengths. 

This paper presents a novel data-driven methodology for discovering and ranking source-impact pathways using random forest regression (RFR) and feature importance that can be applied to ensembles of simulated and/or observed climate data. 
RFR \cite{bishop2006pattern,breiman2001random} is an ensemble learning method, typically used for classification or regression, that operates by constructing an ensemble of decision trees at training time, each of which creates a set of if-then-else rules to approximate a dataset or function \cite{quinlan1986induction,shalev2014understanding}.  RFR has a number of advantages over other data-driven methods, including easy offline training, efficiency, interpretability and built-in feature importance metrics (quantitative measures of how much influence a particular input variable has on the model's output prediction \cite{casalicchio2019visualizing}).  
While RFR and feature importance are both well-known in the field of machine learning (ML), we propose herein a first-of-its-kind combination of these two concepts that enables both source-impact pathway identification and ranking, and operates by: (i) training an individual random forest (RF) regressor for each output, (ii) calculating pairwise feature importance values between the inputs and outputs across all RF regressors, and (iii) converting these feature importances into a pathway network, which can be represented as a directed graph.  This ``outer loop'' algorithm for identifying relationships from data is the primary innovation in this work.  Indeed, in steps (i) and (ii) of the algorithm, RFR and SHAP could easily be swapped with any regression and sensitivity metric, respectively. 

After verifying our methodology on a manufactured problem, we deploy it on our 
targeted climate exemplar problem: the 1991 volcanic eruption of Mount Pinatubo in the Philippines, a stratospheric aerosol injection (SAI) event, the climate impacts of which have been well documented and studied \cite{hansen1992potential,kilian2020impact,parker1996impact,self1993atmospheric,Robock2000, Timmreck2012, Marshall2022}. 
This eruption, which took place on June 15, 1991, was the second largest volcanic eruption of the twentieth century \cite{newhall1997cataclysmic} and injected approximately twenty million tons of sulfur dioxide (SO$_2$) into the stratosphere, causing surface temperatures to decrease for up to two years after the eruption due to a reduction in shortwave radiation, and stratospheric temperatures to increase due to the greenhouse effect of increased sulfates in the stratosphere. The Mount Pinatubo eruption is ideal for feature and pathway detection because of its well-known impacts governed by well-characterized stratospheric dynamics, and the large signal-to-noise ratio in climate variables such as aerosol optical depth (AOD), stratospheric temperature, surface temperature and others.  We utilize as training data an ensemble of simulations of the Mount Pinatubo eruption performed using a modified version of the fully-coupled E3SM version 2 (E3SMv2) \cite{golaz2022doe} known as E3SMv2-Stratospheric Prognostic Aerosols (E3SMv2-SPA) \cite{brown2024validating}, 
augmented to 
simulate prognostically the evolution of aerosols in the stratosphere.  We note that several alternative methods for identifying/detecting pathways within this data set have been proposed during the past 1-2 years, each with its own strengths and weaknesses.  These methods are based on techniques such as changepoint detection \cite{Yarger:2024, Shi-Jun:2024}, echo state networks (ESNs) \cite{Goode:2024, Ries:2024, McClernon:2024}, space-time dynamic models \cite{Garrett:2024}, conditional multi-step fingerprinting \cite{Wentland2024}, operator neural network \cite{Hart:2023, hart2024}, and in-situ profiling \cite{Steyer:2024, Watkins:2024}.  

The remainder of this paper is organized as follows.  In Section \ref{sec:method}, we describe our new RFR and feature importance-based method for source-impact pathway detection. In Section \ref{sec:data}, we describe the two test cases used to develop/verify our method: (i) a set of synthetic coupled equations, 
and (ii) and the actual Mount Pinatubo eruption itself, simulated in E3SMv2-SPA. 
Results for each of these two benchmarks are presented in Section \ref{sec:results}. 
In Section \ref{sec:conclusion}, we provide conclusions and describe possible directions for future work. 


\section{Method}\label{sec:method}

The goal of our method is to use time-series data to create a directed graph comprised of nodes and edges, in which the nodes represent features of interest and the edges represent relationships between these features.  These relationships are directional, weighted and have a time lag associated with them.  Once a graph is fully constructed, it can be used to trace out source-impact pathways, defined as the interactions of a set of variables in space-time due to an external forcing, within the provided time-series data (in our case, climate data).  

Suppose we are given  a 
set of time-series data $\{~F_1, ~F_2, ... ~F_n\}$, where $~F_i \in \mathbb{R}^K$ and $n, K \in \mathbb{N}$.  In this notation\footnote{We note that our workflow allows for the $~F_i$ to have different lengths $K_i$, but assume that these vectors have the same length to facilitate the presentation of the method.}, the $j^{th}$ entry of $~F_i$, denoted by $F_i(j)$, represents the value of $~F_i$ at time $j$ for $1 \leq j \leq K$.  From this point forward, we will refer to $\{~F_i\}$ as the set of ``features" used in our analysis.  In the climate application considered herein, each feature $~F_i$ is a time-series of a variable associated with a given spatial location and outputted by a climate model, e.g., temperature, aerosol optical depth (AOD), etc.  As a concrete example, suppose we are given time-series data for two variables, temperature and AOD, at five spatial locations.  In this case, our features $~F_1, ..., ~F_5$ and $~F_6, ...., ~F_{10}$ represent the  temperature and AOD time-series, respectively, at locations $1, ...,5$, so that $n 
= 10$. In general, if we are given $m$ variables at $N$ spatial locations, for $m, N \in \mathbb{N}$, the total number of features will be $n = mN$.  

\begin{figure}[hbt!]
    \centering
    \includegraphics[width=0.80\linewidth]{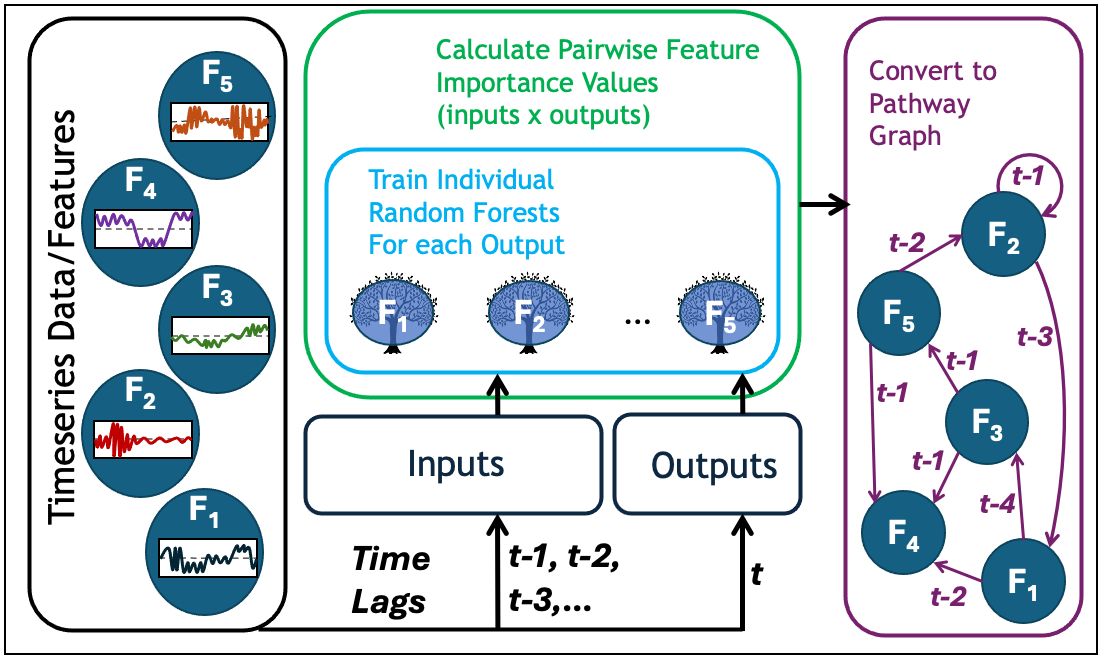}
        \caption{Visual depiction of the RFR-based pathway construction approach described in Section \ref{sec:method} assuming $~F = \tilde{~F}$.} \label{fig:rfr_workflow}
\end{figure}

\begin{figure}[hbt!]
    \centering
    \includegraphics[width=0.90\linewidth]{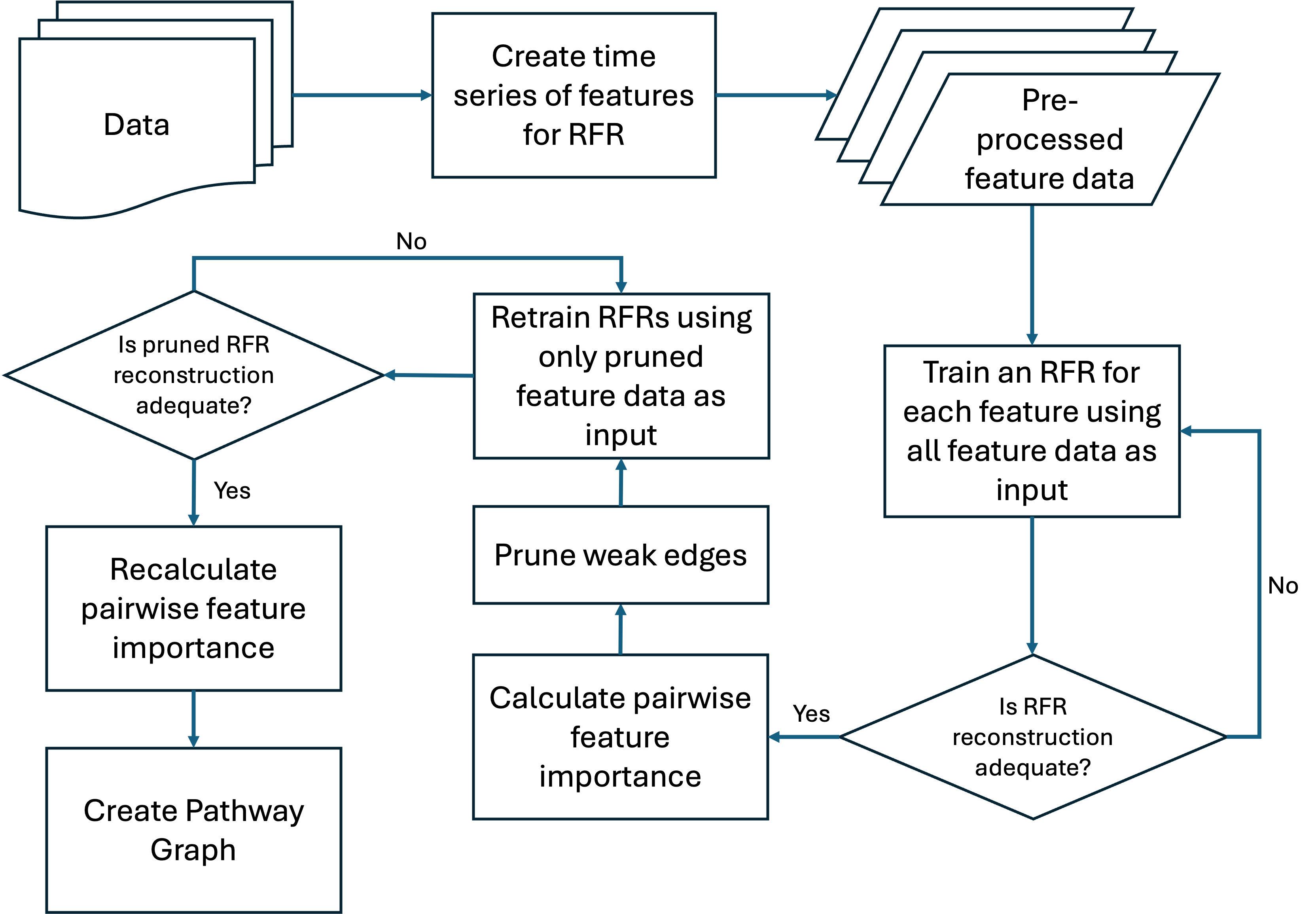}
        \caption{Schematic of RFR and Feature importance-based source-impact pathway construction method described in Algorithm \ref{alg:pseudocode}.}
        \label{fig:rfr_schematic}
\end{figure}

Our pathway detection algorithm consists of several key steps, described in detail below and summarized succinctly in Algorithm \ref{alg:pseudocode} and Figures \ref{fig:rfr_workflow}-- \ref{fig:rfr_schematic}.  Although the discussion herein is focused on RFR and SHAP feature importance as the primary workhorses in Algorithm \ref{alg:pseudocode}, we emphasize that this algorithm is not tied to these approaches.

\subsection{Temporal Lag Selection and Data Pre-Processing} \label{sec:preproc}

The first step in our approach is to choose a set of temporal lags $L:=\{l_1, ..., l_q\}$ (for $l_i, q \in \mathbb{N}$, where $1 \leq i \leq q$) to investigate.  To give a concrete example, if we choose $L$ to be  $L = \{ 1, 3, 5\}$, the features to be investigated are $F_i(t-1)$, $F_i(t-3)$ and $F_i(t-5)$ for $1 < i < n$ and $2<t \leq K $ with $t \in \mathbb{N}$.  
Ideally, the choice of lags $L$ should be informed by the dynamics of the problem and the physical processes of interest.

In developing our method, we found that it is often important to pre-process the data contained in $~F:=\{~F_1, ~F_2, \cdots, ~F_n\}$.  
The most common pre-processing to consider is a spatial dimension reduction of the features in $~F$.  Suppose the data comprising $~F$ are given at a large number of spatial locations, so that $N \gg 1$ and hence $n\gg1$.  
The simplest spatial dimension reduction approach to apply is a basic  averaging of the data in space, achieved by first
decomposing the spatial domain into a set of $\tilde{N} \ll N$ subdomains, e.g., by splitting the grid into latitudinal bands or Intergovernmental Panel on Climate Change (IPCC) regions \cite{IPCC:2020}, then computing a mean of the data in $~F$ over each subdomain.  The result is a new set of data $\tilde{~F}:=\{ \tilde{~F}_1, ..., \tilde{~F}_{\tilde{n}}\}$, where $\tilde{n}=\tilde{N}m$, and $\tilde{~F}_i \in \mathbb{R}^{K}$.  We note that other pre-processing approaches can be beneficial to our analysis, e.g., feature normalization; these are problem specific, and discussed later, in Section \ref{sec:results}.  In Algorithm \ref{alg:pseudocode}, we use the notation $preProc(~F)$ to denote any pre-processing/dimension reduction performed on the data $~F$ prior to applying RFR.



\subsection{Training of Individual RF Regressors to Predict a Given Feature of Interest} \label{sec:train}

RFR \cite{bishop2006pattern, breiman2001random} is a supervised, data-driven technique that utilizes an ensemble of decision trees for regression.  Each decision tree is trained on a subset of the training data in order to reduce overfitting; this is commonly referred to as bagging or bootstrap aggregation.  
Output values predicted by the decision trees are combined within RFR using a majority voting.

The second step in our workflow is to train individual RF regressors to predict  feature $i$,  $\tilde{~F}_i$ (the output), with the set of all features at all times, $\tilde{~F}$, as inputs.
To do this, for each output $\tilde{~F}_i$ with $1 \leq i \leq \tilde{n}$, we create an input vector by concatenating all possible features at all possible times $t-l_j$ for all possible lags $l_j \in L$,     
%
 and train a RF regressor to predict $\tilde{~F}_i$, as described in Algorithm \ref{alg:pseudocode}.  
In Algorithm \ref{alg:pseudocode}, the subroutine $trainRFR(Inputs, Outputs)$ refers to the RF training step of our workflow.  
In our current implementation, we perform this step by utilizing Python's sci-kit learn package \cite{SciKitLearn:2011}. 
The number of decision trees and decision tree depth are controlled parameters in the RFR and hence 
input parameters for sci-kit learn.  For the results presented herein, we used 100 trees, each with a depth of four.

\subsection{Evaluation of RF Regressors' Goodness of Fit} \label{sec:evaluate}

Once our RF regressors are trained, the next step is to assess their goodness of fit using appropriate evaluation metrics.  
Here, we rely on two commonly-used evaluation metrics: $R^2_{adj}$ (the adjusted coefficient of determination) and $RMSE$ (the root mean square error), defined as  
\begin{equation}
    R^2_{adj} = 1 - \left[\frac{(1-R^2)(n-1)}{n-p-1}\right]
\label{eq:r2_adj}
\end{equation}
where
\begin{equation}
    R^2 = 1-\frac{\sum_{i=1}^n{(y_i - \hat{y_i})^2}}{\sum_{i=1}^n{(y_i - \bar{y})^2}}
\label{eq:r2}
\end{equation}
and 
\begin{equation}
    RMSE = \sqrt{\frac{\sum_{i=1}^n{(y_i - \hat{y_i})^2}}{n}}. 
\label{eq:rmse}
\end{equation}
In Equations \eqref{eq:r2_adj}--\eqref{eq:rmse}, $y_i$ is a reference data point (usually a ground truth measurement, in our case, it is either the synthetic or climate data we are training to), $\hat{y_i}$ is the corresponding RFR prediction, $\bar{y}$ is the mean of the reference data, $n$ is the number of samples,  and $p$ is the number of independent variables. We prefer in general $R^2_{adj}$ over $R^2$, the coefficient of determination, as $R^2_{adj}$ tends to be less sensitive to the number of inputs.  We generally look for $R^2_{adj} > 0.75$ and an $RMSE$ on the order of at most 0.15.
If these tolerances are not met, it may be necessary to redo the training step of the workflow with different data, pre-processing and/or lags.

\subsection{Feature Importance Calculation} \label{sec:fi}

Feature importances (also referred to herein as ``weights" and denoted by $w_i$) are obtained for machine learning models like RFR via a post-processing, 
and are computed herein  
using Python's scikit-learn package \cite{SciKitLearn:2011}. 
The relative magnitudes of these feature importances are representative of the relative strength of the relationship between any two features. 
In the present analysis, we utilize the SHapley Additive exPlanation (SHAP) method to calculate the feature importances/weights between each input feature and output feature $(\tilde{F}_i(t), \tilde{F}_j(t-l))$ \cite{lundberg2017unified}.
This approach draws on Shapley values and cooperative game theory \cite{shapley1953value}, which  aims to evaluate how important any given player is to winning a cooperative game.  We note that SHAP is just one possible feature importance measure; other commonly used feature importances include Gini \cite{Gini, Nichol:2021}, permutation \cite{PFI}, amongst others. We select the SHAP approach as several recent studies have shown that SHAP values can achieve better results for assessing feature importance in machine learning models than other metrics for various problems of interest 
\cite{marcilio2020explanations,wang2024feature,dunn2021comparing,kumar2020problems}. Additionally and importantly, SHAP proved to be a computationally 
efficient metric for our analysis. 



\subsection{Extension to Data Ensembles} \label{sec:ensembles}

Our algorithm description thus far has assumed that the input data consist of a single set of features $~F$.  However, since climate analyses typically operate on ensembles of data, generated, for example by  slightly perturbing the initial condition,  there is a need to have a way of applying our workflow to ensembles of features $\{\{~F\}^r\}_{r=1}^R$ for $R \in \mathbb{N}$.  The multiple ensemble member case is handled by repeating the RFR training process described above and in Algorithm \ref{alg:pseudocode} for each set of ensemble members $\{~F\}^r$ and averaging the SHAP feature importance weights across the ensemble members for each pairwise input to output.  In the case ensembles are used, all goodness of fit metrics (Section \ref{sec:evaluate}) are calculated on a per ensemble basis.

\subsection{Pruning of Edges and Construction of Directed Pathway Graph}  \label{sec:prune}

The final step in our procedure involves constructing a directed pathways graph from the RF regressors and feature importance information. 
For each pair of features $\tilde{~F}_i$ and $\tilde{~F}_j$, we draw a set of edges.  The weight of each edge is given 
by the value of the feature importance.
After calculating the mean and standard deviation of the SHAP weights across a set of ensembles (Section \ref{sec:ensembles}), we use these values to prune potential edges 
in our pathway graph 
using the following criteria:
\begin{enumerate}
    \item We select the top four incoming edges, i.e., the four largest $w_i$, to each feature.  
    \item We prune edges with $w_i$ that does not exceed a minimum threshold $\delta$. In all of our numerical results (Section \ref{sec:results}), we used a threshold of $\delta  = 1.0 \times 10^{-4}$.  
    \item We prune edges where the standard deviation ($\sigma$) of $w_i$ exceeds the mean weight.
    \item We prune edges that are not represented in a majority of ensemble members (a discussion of how we treat 
    an ensemble of data is given in Section \ref{sec:ensembles}).
\end{enumerate}
The first two criteria enumerated above are data specific, and may need to be refined on a dataset by dataset basis.

Recall that, in our feature importance-based workflow, each RF regressor predicts one variable using all features as inputs, and one SHAP value is assigned for each incoming edge. As described in Section \ref{sec:train}, our feature importance-based analysis is repeated many times, iterating through all variables of interest. While Step 1 in the procedure outlined above discusses only incoming edges in our directed pathway graph, the nodes of this graph will have both incoming and outgoing edges.  The outgoing edges and their associated SHAP values can then be inferred from the incoming edges, as they were incoming edges for alternate nodes in the graph. When we combine all these edges from all the individual calculations, we obtain a graph whose nodes have both incoming and outgoing edges. 

 Our pruning methodology, which effectively sets all the pruned weights to $w_i = 0$, is referred to as $prune(Edges)$ in Algorithm \ref{alg:pseudocode}.
Once pruning has been completed, we again use the goodness of fit metrics described in Section \ref{sec:evaluate} to 
evaluate the RFR reconstructions of the features.  If reconstruction is poor, refinement of the pruning criteria may be needed.

Once the pruned RFR reconstruction has been determined to be a good fit to the original data, we translate our feature importance 
values $w_i$ into a 
  directed graph, denoted by the $createPathwaysGraph(Nodes, Edges)$ routine in Algorithm \ref{alg:pseudocode}.  Here, each node represents a feature $\tilde{~F}_i$, and each edge between node $\tilde{~F}_i$ and node $\tilde{~F}_j$ is weighted by the feature importance $w_i$ corresponding to the relevant pair provided $w_i > 0$ (see Figure \ref{fig:rfr_workflow}, right column).  To make graph depiction more manageable, temporal lags at which relationships between features have been detected are represented by numbers that are printed over each edge (see Figure \ref{fig:rfr_workflow}, right column).   Once a pathway graph is fully constructed, pathways can be inferred by following the directed edges in the graph from one feature to another. 




\begin{algorithm}
\caption{Pseudocode to create pathway graph from a single realization of spatio-temporal data.  The algorithm is depicted visually as a workflow in Figure \ref{fig:rfr_workflow}.  If an ``Abort" is hit, it is necessary to rerun the algorithm with different features, pre-processing, temporal lags and/or pruning criteria 
 }\label{alg:pseudocode}
\begin{algorithmic}
\Require Set of spatio-temporal features $~F \in \mathbb{R}^{K \times n}$.
\Comment{Input time-series data}
\State Specify temporal lags $L=\{l_1, ..., l_q\}$.  \Comment{Select temporal lags to be investigated (Section \ref{sec:preproc}).}
\State $\tilde{~F} =preProc(~F) \in \mathbb{R}^{K \times \tilde{n}}.$ \Comment{Apply pre-processing/dimension reduction to $~F$ (Section \ref{sec:preproc})}
\State $Edges \gets \{\}$, $Nodes \gets \{\tilde{~F}_1,\tilde{~F}_2,...,\tilde{~F}_n,\}$.   \Comment{Initialize Edges and Nodes arrays}
\For {$i$ in $\{1, ..., \tilde{n}\}$}
       \State $Outputs = \{\tilde{~F}_i\}$, $Inputs = \{\}$.  \Comment{Create a set of Outputs and Inputs, and populate}
       \For {$j$ in $\{1, ..., \tilde{n}\}$}
         \For {$k$ in $\{ q+1, ..., K\}$}
         \For {$l$ in $\{1, ..., q\}$}
            \State Append $F_j(k-l)$ to $Inputs$.  
         \EndFor
         \EndFor
       \EndFor
    \State $RFR^{i} \gets trainRFR(Inputs,Outputs)$
    \Comment{Train RFR to predict $\tilde{~F}_i$ (Section \ref{sec:train})}
    \State Calculate $R^2_{adj}$ and $RMSE$.  \Comment{Calculate goodness of fit metrics (Section \ref{sec:evaluate})}
    \If{$R^2_{adj}$ sufficiently large and $RMSE$ sufficiently small}
    \State $Edges^{i} \gets featureImportance(RFR^{i})$
    \Comment{Find edge weights going into $\tilde{~F}_i$ (Section \ref{sec:fi})}
    \State Concatenate $Edges^{i}$ to $Edges$
      \EndIf
\State $prune(Edges)$
\Comment{Prune out weak or irrelevant edges (Section  \ref{sec:prune})}  
\State Calculate $R^2_{adj}$ and $RMSE$.  \Comment{Calculate goodness of fit metrics (Section \ref{sec:evaluate})}
 \If{$R^2_{adj}$ sufficiently large and $RMSE$ sufficiently small}
\State $Graph = createPathwayGraph(Nodes,Edges)$ \Comment{See Section \ref{sec:prune}}
\Else 
\State Abort.  
\EndIf
\State Abort.
\EndFor
\end{algorithmic}
\end{algorithm}

\section{Data}\label{sec:data}


Since we are proposing a novel data-driven approach for creating pathway graphs, 
it is important to take a systematic approach to method verification and validation.  
Herein, we adopt a tiered verification approach that applies Algorithm \ref{alg:pseudocode} to two increasingly-complex benchmarks, described in more detail below.   
First, we develop a system of synthetic coupled equations, where the relationships between the variables are explicitly given and can be controlled to test potential corner cases.  
Next, we apply our method to data obtained by simulating the 1991 Mount Pinatubo eruption in the Philippines using the fully coupled E3SMv2-SPA \cite{brown2024validating} (described in more detail below).
While some variable relationships and pathways in the second benchmark are well-understood, making it a good verification test case, the underlying physical processes are very complex, 
leaving room for our model to discover relationships beyond those that are expected. 

\subsection{Synthetic Coupled Equations}\label{sec:synthetic_res}

For our first benchmark, we consider a five member ensemble of data generated by simulating the following analytically-given time-series of coupled equations:  

\begin{equation} 
\label{eq:synthetic}
    \begin{aligned}
    W_t & = 0.9W_{t-1} + \epsilon_{W_t} \\
    X_t & = 0.8X_{t-1} + 0.5W_{t-1} + \epsilon_{X_t} \\
    Y_t & = -0.9W_{t-1} + \epsilon_{Y_t} \\
    Z_t & = 0.6X_{t-1} + 0.5Y_{t-1} + \epsilon_{Z_t}.
    \end{aligned}
\end{equation}
In Equation set \eqref{eq:synthetic}, $W_t$, $X_t$, $Y_t$ and $Z_t$ are the primary variables/features evaluated at time $t$, and $\epsilon_{W_t}$, $\epsilon_{X_t}$, $\epsilon_{Y_t}$ and $\epsilon_{Z_t}$ are random variables representing background noise, herein sampled separately from a uniform $[-0.5,0.5]$ distribution.  We use \eqref{eq:synthetic} to create a five member ensemble, each containing 750 time-values for each variable.    
Each time-series is initialized with a random seed based on the noise term for the variable.
Different ensemble members are generated by varying the noise terms $\epsilon_{W_t}$, $\epsilon_{X_t}$, $\epsilon_{Y_t}$ and $\epsilon_{Z_t}$.  Since this synthetic dataset has known, clearly-encoded variable interactions, it enables us to test our method's ability to identify these known relationships before proceeding to more sophisticated test cases.  Specifically, we expect the variable $W_t$ to be most strongly dependent on $W_{t-1}$ (defining a pathway from $W_{t-1}$ to $W_t$),  the variable $X_t$ to be most strongly dependent on $X_{t-1}$ followed by $W_{t-1}$ (defining pathways from $X_{t-1}$ and $W_{t-1}$ to $X_t$), etc.  It is noted that, since the synthetic equations \eqref{eq:synthetic} only contain a temporal dependence, it is not necessary to perform a spatial dimension reduction (Section \ref{sec:preproc}) prior to applying our method.

\subsection{Mount Pinatubo Eruption Simulated Using E3SMv2-SPA}\label{sec:e3sm_res}
Our second benchmark involves analyses on data from simulations of the 1991 eruption of Mount Pinatubo using a modified version of E3SMv2.  
Typically, ESMs such as E3SMv2 prescribe the location, absorption, and scattering of stratospheric aerosol from explosive volcanic eruptions (i.e., take them from an input file), rather than simulating them explicitly within the model.
In order to study the impacts of the volcanic eruption  using the fully-coupled E3SMv2, 
we extended this model to handle stratospheric aerosols prognostically, that is, to enable us to 
simulate the dynamic evolution of volcanic sulfate from an injection of SO${_2}$, together with downstream climate impacts.  
We will refer to the resulting fully-coupled version of E3SMv2, described in more detail in \cite{brown2024validating}, as E3SMv2 Stratospheric Prognostic Aerosol, or E3SMv2-SPA\footnote{Available at: \url{https://github.com/sandialabs/CLDERA-E3SM}.}.  

\begin{figure}[htbp!]
        \begin{center}
                \subfigure[1 day]{
      \includegraphics[width=0.48\textwidth]
                      {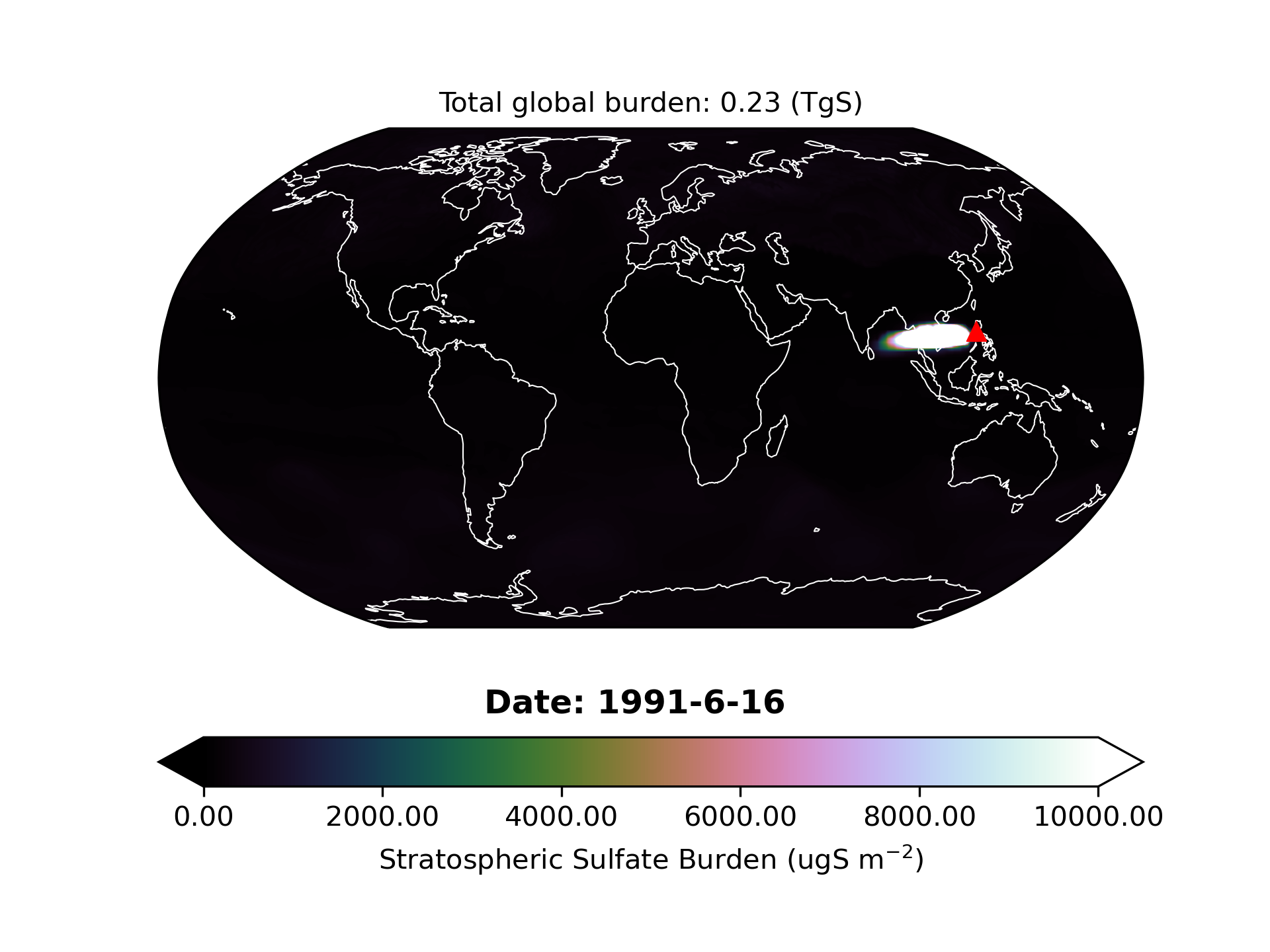}}
                           \subfigure[6 days]{
      \includegraphics[width=0.48\textwidth]
                      {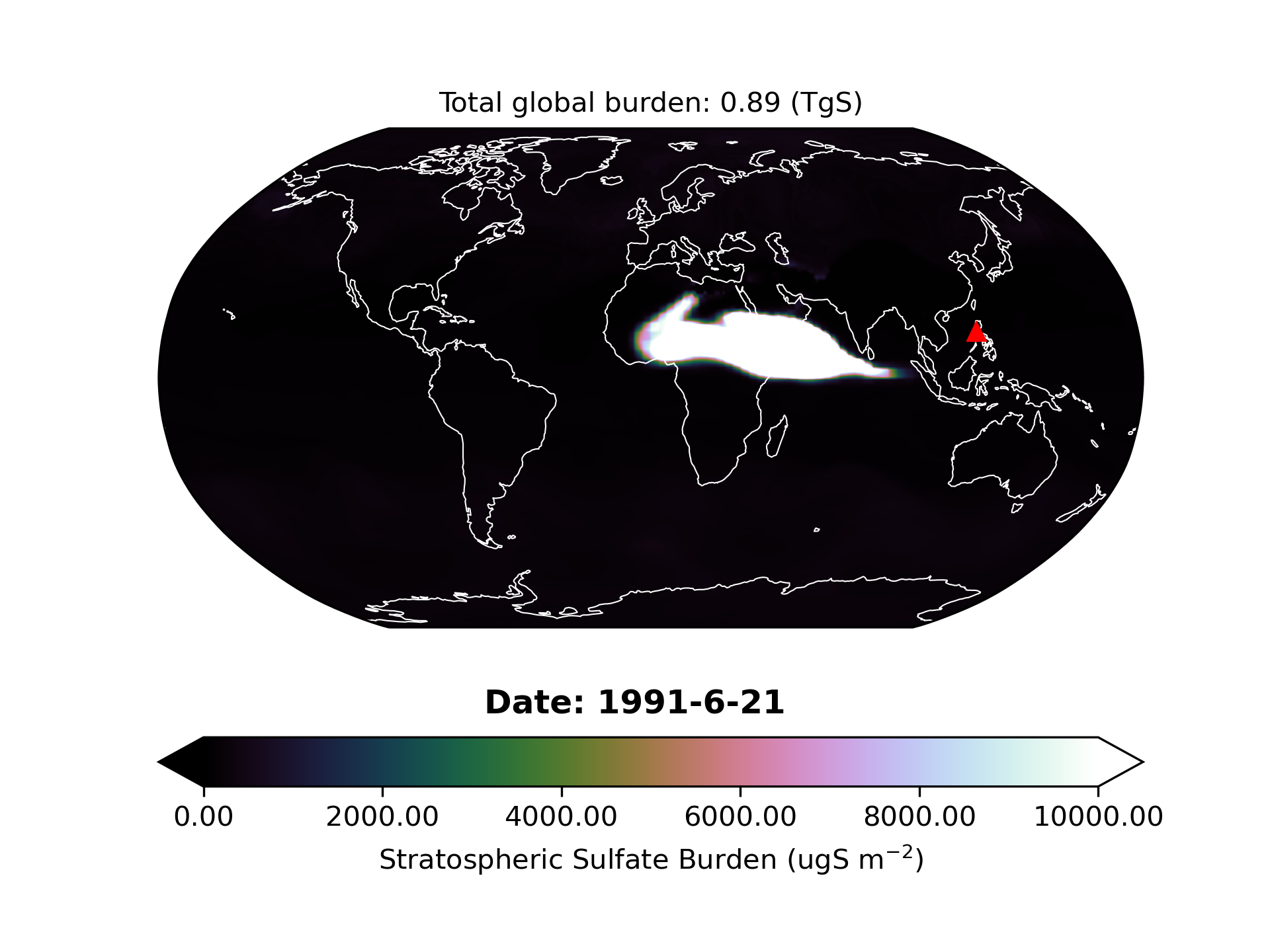}}
                      \subfigure[11 days]{
        \includegraphics[width=0.48\textwidth]
                      {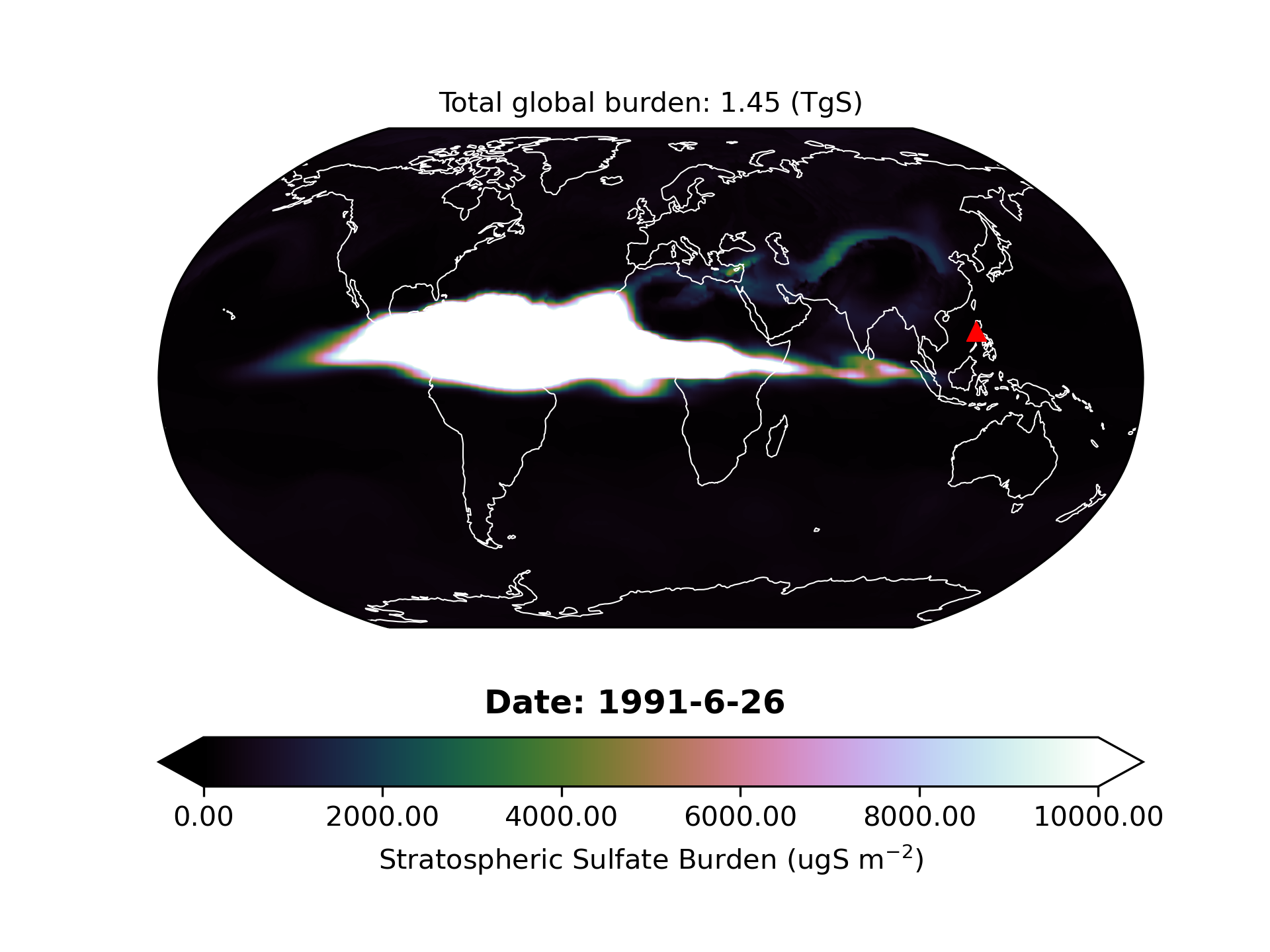}}
                       \subfigure[16 days]{
        \includegraphics[width=0.48\textwidth]
                      {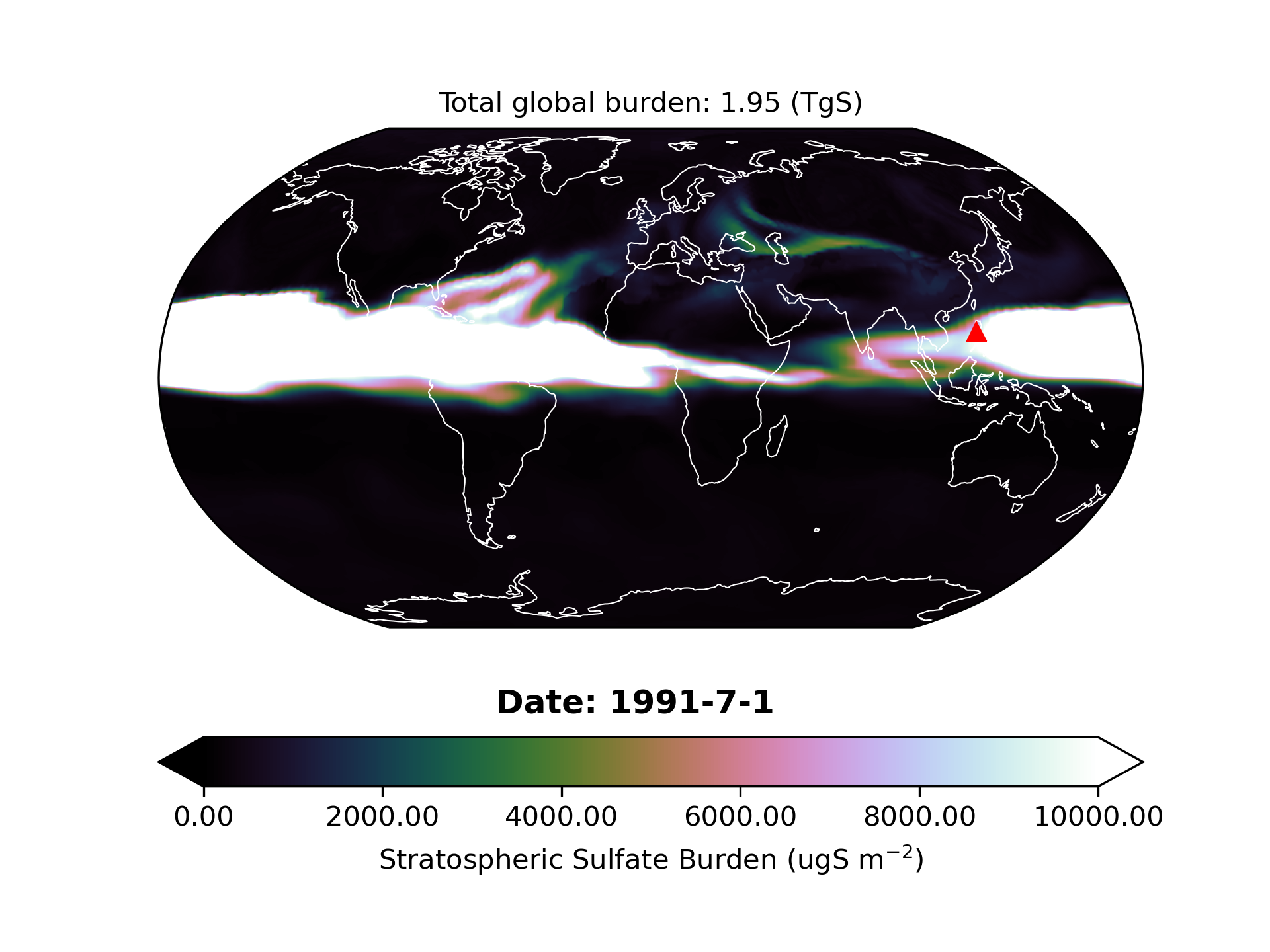}}
                       \subfigure[21 days]{
        \includegraphics[width=0.48\textwidth]
                      {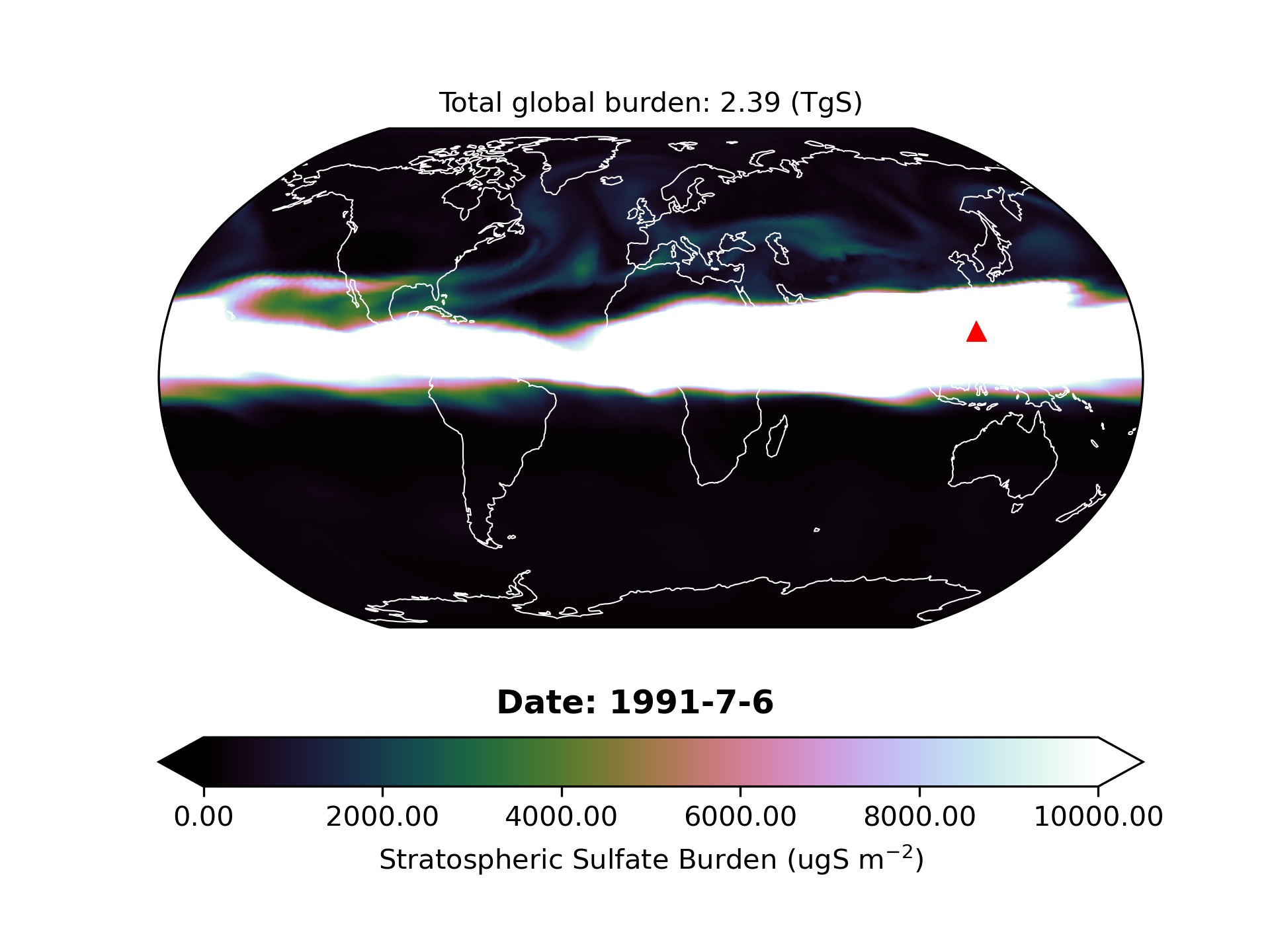}}
                        \subfigure[31 days]{
        \includegraphics[width=0.48\textwidth]
                      {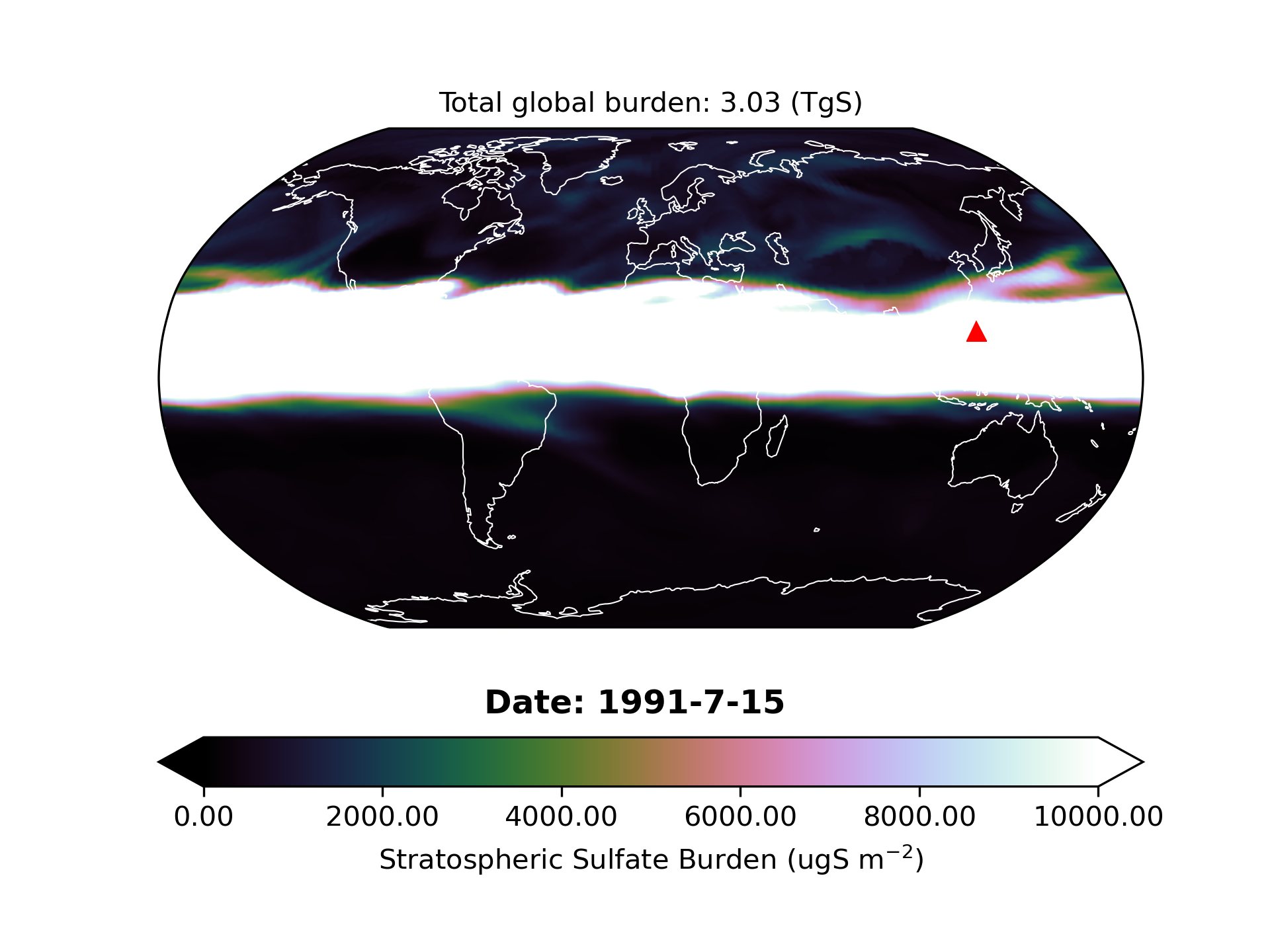}}
        \end{center}
	\caption{The stratospheric sulfate burden at different days up to roughly one month following the Mount Pinatubo eruption.  The location of Mount Pinatubo is marked with a red triangle.  The sulfates have encircled the Earth by approximately 21 days post eruption (e), and have made their way into the subtropics by approximately 31 days post eruption (f).}
        \label{fig:stratospheric_burden}
\end{figure}

In the stratosphere, we expect to find a pathway in which the formulation of sulfates increases aerosol optical depth (AEROD\_v), which, in turn, increases longwave radiation absorption (FLNT) and leads to an increase in the stratospheric temperature at 50 hPa (T050) \cite{Labitzke1992}. At the Earth's surface, we still expect to see an increase in aerosol optical depth (AEROD\_v) due to the presence of sulfates; however, as a consequence, we expect to see a decrease in the amount of shortwave radiation reaching the surface (FSDS), followed by a lowering of the temperature at the surface (TREFHT) \cite{parker1996impact, Soden2002}.  It is well-known that the aerosol cloud from Mount Pinatubo's June 15, 1991 eruption encircled the globe in just 22 days and filled the entire tropical belt in approximately two months, before spreading to higher latitudes \cite{Polvani:2019}.  This result is confirmed by our simulations of the Mount Pinatubo eruption using E3SMv2-SPA.  Figure \ref{fig:stratospheric_burden} shows the stratospheric sulfate burden 1, 6, 11, 16, 21 and 31 days after the eruption.  The location of Mount Pinatubo is marked with a red triangle.  The global surface temperature was reduced by $-0.5^{\circ}$C by September 1992 \cite{Dutton:1992}. 


In our analysis herein, we focus our attention on five variables, summarized in 
Table \ref{tab:e3sm_vars}.  The reader can observe that we have chosen to utilize the clear-sky versions of the longwave and shortwave radiation, denoted by FLNTC and FSDSC, respectively.  We choose these variables because they provide a clearer signal than FLNT and FSDS.


\begin{table}[hbt!]
    \centering
     \caption{Mount Pinatubo temperature pathway variables}
    \begin{tabular}{cccc}
    \hline \hline
     \textbf{Variable Name}& \textbf{Long Name} &\textbf{Units} & \textbf{Description}\\
     \hline
         \multirow{3}{*}{AEROD\_v} & \multirow{3}{*}{Aerosol Optical Depth} & \multirow{3}{*}{$-$} & 
         column-integrated aerosol \\
         &&&optical depth (missing \\
         &&&data in polar winter)\\
         \hline\multirow{3}{*}{FLNTC} & Clear-sky net longwave & \multirow{3}{*}{W m$^{-2}$} & dominant radiative \\
         & radiative flux at the  && flux in the\\
         & top of atmosphere&&  stratosphere\\\hline
         T050 & Temperature at 50 hPa & K & 
         stratospheric temperature\\\hline
         \multirow{3}{*}{FSDSC} & Clear-sky downward & \multirow{3}{*}{W m$^{-2}$} &
         dominant radiative \\ 
         & shortwave radiative flux  &&flux at the \\
         &at the surface&& surface\\\hline
         TREFHT & Temperature at 2m & K & 
         near-surface air temperature\\\hline
    \end{tabular}  \label{tab:e3sm_vars}
\end{table}

Our analysis was based on five
limited variability (LV) simulations, 
described in more detail in \cite{Ehrmann:2024} and summarized briefly below. 
Limited variability was achieved by matching historical conditions for the Quasi-Biennial Oscillation (an alternating equatorial zonal wind pattern in the stratosphere) and the El Ni\~{n}o Southern Oscillation (changes in tropical Pacific sea surface temperatures).  These two major modes of natural variability respectively precondition the direction of travel of the injected SO$_2$ gas as well as temperature and precipitation values globally \cite{Davey2014}.  The five ensemble members were generated by perturbing the temperature initial conditions on June 1, 1991 with the eruption introducing a point injection of 10 Tg SO$_2$ at 15.15$^{\circ}$N and $120.35^{\circ}$E at an altitude of 18-20km occurring on June 15, 1991.  Simulations ran through December 31, 1998.  For comparison purposes, we also had at our disposal a paired five member set of ``counterfactual" ensembles, containing no volcanic eruption but using the same initial conditions as the ensembles containing the eruption. Simulations   
were performed using a $1^{\circ}$ configuration of E3SMv2-SPA,  which has a resolution of 110 km with 72 vertical layers for the atmosphere, and a resolution of 165 km for land.  
Our RF regressors were trained with daily averaged climate data.  The details of how these data were pre-processed are given later, in Section \ref{sec:results}.  

\section{Results}\label{sec:results}

In the following sections,  we present results for each tier in our tiered verification strategy.
We start with the synthetic data example, where the features are time-series from the system of coupled equations given in \eqref{eq:synthetic} (Section \ref{sec:synth_results}). We then 
apply our method to an ensemble of Mount Pinatubo simulations performed using the fully-coupled E3SMv2-SPA \cite{brown2024validating} (Section \ref{sec:pinatubo-resuls}).






\subsection{Synthetic Coupled Equations} \label{sec:synth_results}

Since the coupled synthetic equations \eqref{eq:synthetic} are relatively simple and well-behaved, and do not have a spatial dependence, no pre-processing was performed on these data. The time lags used in our analysis were $L = \{1, 2, 3, 4, 5\}$. 
While the ``ground truth" only contains lags of 1 time step, we wanted to verify that the algorithm could identify the ``correct" time lag in the presence of extra time lags.

\begin{figure}[hbt!]
    \centering
    \includegraphics[width=0.9\linewidth]{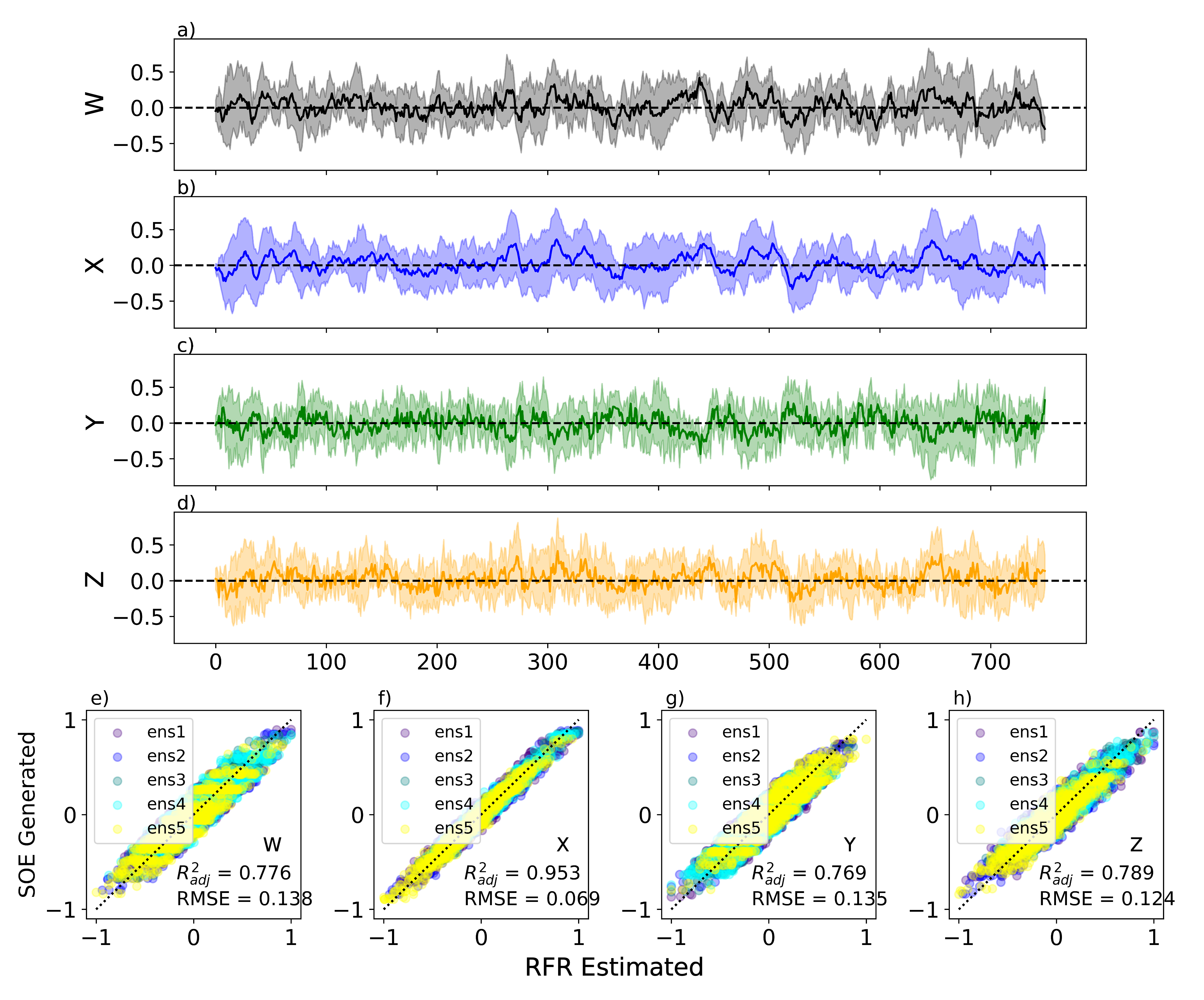}
    \caption{Coupled synthetic equations: ensemble mean goodness of fit statistics. Subfigures a)-d) show the ensemble mean and standard deviation of $W$, $X$, $Y$, and $Z$ respectively. Subfigures e)-h) show the scatterplot comparison between the SOE generated time series and their RFR reconstructions, for $W$, $X$, $Y$,and $Z$ respectively.}
    \label{fig:SOE_GOF}
\end{figure}

After pre-processing the data, we begin by evaluating our RF regressors' ability to reconstruct each feature of interest, which, for the synthetic coupled equation, is each variable $W$, $X$, $Y$ and $Z$.
In Figure \ref{fig:SOE_GOF}, we show the comparison between the time-series of the system of synthetic equations and the RFR reconstruction for each feature. 
Figure \ref{fig:SOE_GOF} subfigures a)-d) show the ensemble mean time series of the data, with the standard deviation shown in the shaded area. Subfigures e)-h) show the scatterplots between the system of equation (SOE) generated data on the $y$-axis and the RFR reconstructions on the $x$-axis, with each ensemble shown in a different color. 
The reader can observe that the RF regressors capture the average signals well, and that the less noisy variables ($X$) are better captured than the more noisy variables ($W$, $Y$, $Z$).  These results are confirmed by Table \ref{tab:SOE_GOF}, which shows the corresponding average coefficient of determination ($R^2_{adj}$) for each feature across all the ensemble members.  The RF regressors are able to fit all the synthetic features well ($R^2_{adj}>0.75$ and $RMSE < 0.15$).  


\begin{table}[hbt!]
    \centering
     \caption{Coupled synthetic equations: goodness of fit statistics of the RFR reconstruction.}
    \begin{tabular}{ccccc}
         \hline \hline
         \textbf{Variable}&\textbf{$R^2_{adj}$ mean} & \textbf{$R^2_{adj}$ $\sigma$}&\textbf{$RMSE$ mean} & \textbf{$RMSE$ $\sigma$}\\ 
        \hline
        $W$&0.776&0.043&0.138&0.009\\
        $X$&0.953&0.010&0.069&0.007\\
        $Y$&0.769&0.036&0.135&0.007\\
        $Z$&0.789&0.040&0.124&0.012\\
         \hline
    \end{tabular}
    \label{tab:SOE_GOF}
\end{table}

\begin{figure}[hbt!]
    \noindent
    \centering
    \includegraphics[width=0.45\linewidth]{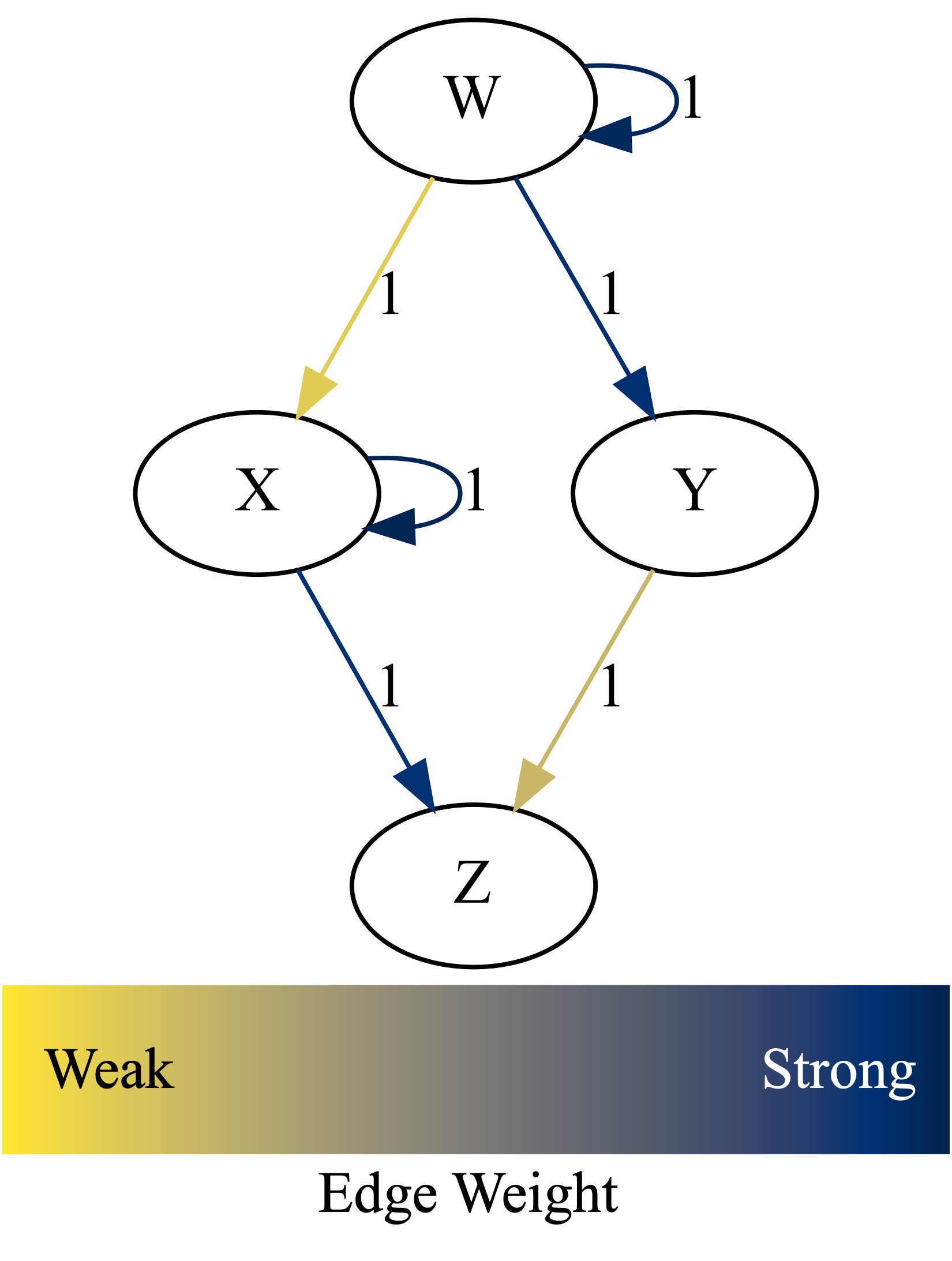}
    \caption{Coupled synthetic equations: pathway graph. Features are shown in circles, feature connections are shown by the arrows pointing from source feature to target feature. Numbers next to the arrows are the time lags associated with the connection. The edge colors represent SHAP weights, blue indicating higher values and yellow indicating lower values.}
    \label{fig:SOE_graph}
\end{figure}

Next, we calculate and report the feature importances for the coupled synthetic data set, which are summarized in Table \ref{tab:SOE_results}, sorted from highest to lowest feature importance (weight). In this table, the variables in the  ``Target" column are the variables being predicted, $W_t$, $X_t$, $Y_t$ and $Z_t$, and the variables in the  ``Source" column are the dependent variables  $W_{t-l}$, $X_{t-l}$, $Y_{t-l}$ and $Z_{t-l}$, where the value of $l$ is given by ``Lag" variable.  
The reader can observe that the ordering of variable dependencies in Table \ref{tab:SOE_results} is consistent with the known relationship encoded in the governing equations \eqref{eq:synthetic}, e.g., the variable $X$ depends on itself and $W$ at the previous time-step, the variable $Z$ depends on the variables $X$ and $Y$ at the previous time-step, etc.
It is important to recognize that our RFR and feature importance approach cannot recover the actual coefficients pre-multiplying each term on the right-hand side of \eqref{eq:synthetic}; however, the approach \textit{is} able to detect correctly the relative strength of dependence for each variable and each time lag.  
Figure \ref{fig:SOE_graph} is a conversion of Table \ref{tab:SOE_results} into its corresponding pathway directed graph form.    No edge pruning was performed to generate this graph.
Each feature is shown in a labeled circle with arrows indicating the strength and the direction of the feature connections from source to target feature. The numbers labeling the connection arrows denote the time lag associated with the connection.

\begin{table}[hbt!]
    \centering
     \caption{Coupled synthetic equations: pathway edge weights and their standard deviations. }
    \begin{tabular}{ccccc}
         \hline \hline
         \textbf{Source}&\textbf{Target}&\textbf{Lag (time step)}&\textbf{SHAP Weight}& \textbf{Weight $\sigma$}  \\ 
         \hline
        $X$&$X$&1&0.243&0.014\\
        $X$&$Z$&1&0.225&0.014\\
        $W$&$Y$&1&0.229&0.010\\
        $W$&$W$&1&0.240&0.016\\
        $Y$&$Z$&1&0.051&0.009\\
        $W$&$X$&1&0.027&0.004\\
        \hline
    \end{tabular}
    \label{tab:SOE_results}
\end{table}

\subsection{Mount Pinatubo Eruption} \label{sec:pinatubo-resuls}



For our Mount Pinatubo exemplar,
we worked with average daily data for each variable of interest (defined earlier in Table \ref{tab:e3sm_vars}). 
The Mount Pinatubo simulations were 2770 days long (1 June 1991 to 31 December 1998) and used time lags of $L = \{ 1,6,11,16,...,61\}$. 
Since the Mount Pinatubo data are defined on climate grids having a   $1^{\circ}$ resolution spatial grid, spatial dimension reduction was needed to make our workflow and subsequent analysis/interpretation tractable.  Toward this effect, for each variable of interest from each ensemble member, we reduced the spatial dimension by taking a regional average.  
We considered two possible regional averages: (i) an average over the entire globe, and (ii) a regional average over seven latitudinal (or zonal) bands, summarized in Table \ref{tab:lat_bands}.  
We additionally performed some normalization of the input variables to get all the variables on a similar scale.  
Specifically, we subtracted from each ensemble member its corresponding paired counterfactual ensemble member, so as to isolate impacts from the eruption, and normalized the resulting differenced time-series to be between $-1$ and 1.

\begin{table}[hbt!]
    \centering
    \caption{Latitudinal zones}
\begin{tabular}{cc}
     \hline \hline
     \textbf{Zone}& \textbf{Extent}\\
    \hline
     Polar North&66.5$^{\circ}$N - 90$^{\circ}$N\\
     Temperate North&35$^{\circ}$N - 66.5$^{\circ}$N\\
     Subtropical North&23.5$^{\circ}$N - 35$^{\circ}$N\\
     Tropical&23.5$^{\circ}$S - 23.5$^{\circ}$N\\
     Subtropical South&35$^{\circ}$S - 23.5$^{\circ}$S\\
     Temperate South&66.5$^{\circ}$S - 35$^{\circ}$S\\
     Polar South&90$^{\circ}$S - 66.5$^{\circ}$S\\
     \hline
\end{tabular}

\label{tab:lat_bands}
\end{table}

We will consider the two known pathways 
of interest described in Section \ref{sec:e3sm_res}, (i) the stratospheric warming pathway (AEROD\_v $\to$ FLNTC $\to$ T050), and (ii) the surface cooling pathway (AEROD\_v $\to$ FSDSC $\to$ TREFHT), which are treated 
independently of one another in our analysis.

\subsubsection{Spatial Dimension Reduction via Global Averaging} \label{sec:e3sm_global_avg}

\noindent {\bf \textit{Stratospheric Warming Pathway}} \\ 

\begin{figure}[hbt!]
    \centering
    \includegraphics[width=0.9\linewidth]{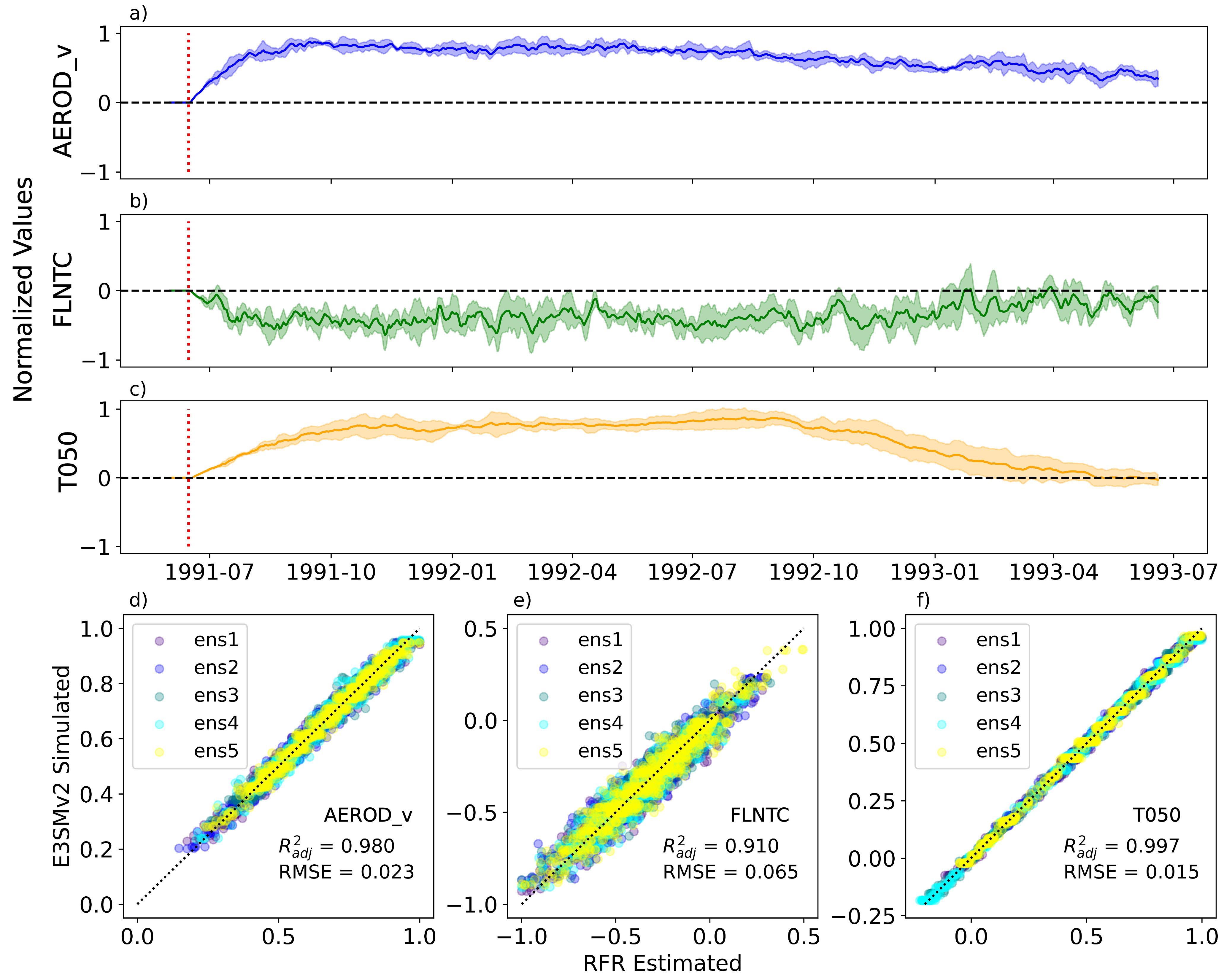}
    \caption{Mount Pinatubo exemplar: goodness of fit statistics for the stratospheric warming pathway using globally-averaged variables. Subplots a), b) and c) show the input time-series data simulated using E3SMv2-SPA for the variables AEROD\_v, FLNTC and T050, respectively. Shaded areas show the ensemble standard deviation.  The dark line is the ensemble mean.  The dashed red line shows the date of Mount Pinatubo eruption. Subplots d), e), and f) show the scatterplots between the normalized features ($y$-axis) and their RFR reconstructions ($x$-axis).
    }
    \label{fig:e3sm_strat_GOF}
\end{figure}

We first consider the simplest spatial dimension reduction, obtained by taking a global average of each of the  variables of interest.  
As before, we first evaluate our RF regressors' skill at reconstructing the features of interest relevant to the stratospheric warming pathway, AEROD\_v, FLNTC and T050. 
We start by calculating the goodness of fit metrics $R_{adj}^2$ and $RMSE$ (see Section \ref{sec:evaluate}) for the three variables defining the stratospheric warming pathway, 
AEROD\_v $\to$ FLNTC $\to$ T050. 
These values are reported in Table \ref{tab:e3sm_strat_GOF}.   The reader can observe that 
$R^2_{avg}$ is above 0.9 for all five variables with an $RMSE$ of at most 6.5\%.  
Additionally, the standard deviation is negligible for both quantities. 
Corroborating these results are the images provided in Figure \ref{fig:e3sm_strat_GOF}, which shows the time-series of the three variables considered AEROD\_v, FLNTC, and T050 (subplots a), b) and c), respectively) and scatterplots between the normalized time-series data and their RFR reconstructions (subplots d), e), f), respectively).  
The reader can observe that FLNTC is the noisiest field, which translates to the lowest $R^2_{adj}$ value and the highest $RMSE$ value for the RFR reconstruction.
The reader will also notice that, although the climate simulation data available ran from June 1, 1991 to December 31, 1998, in our analysis we only consider the first 750 days (approximately 2 years, or June 1, 1991 to June 20, 1993) around the eruption. This is for two reasons: (i) after approximately 2 years, the anomaly calculated by subtracting the counterfactual simulations from the fully coupled Mount Pinatubo simulations returns to zero, and (ii) when we consider the full simulation, only the autocorrelated edges survive our pruning criteria, resulting in a null pathway.

\begin{table}[hbt!]
    \centering
     \caption{Mount Pinatubo exemplar: goodness of fit statistics for the pruned globally-averaged data and the stratospheric warming pathway shown in Figure \ref{fig:e3sm_LV_ens_edges_strat}.}
    \begin{tabular}{cccccc}
    \hline \hline
         \textbf{Variable}&\textbf{$\textbf{R}^2_{adj}$ mean}& \textbf{$\textbf{R}^2_{adj}$ $\sigma$} & \textbf{$RMSE$ mean} & \textbf{$RMSE$ $\sigma$}\\
         \hline
        AEROD\_v&0.980&0.003&0.023&0.001\\
        FLNTC&0.910&0.012&0.065&0.001\\
        T050&0.997&0.001&0.015&0.001\\
         \hline
    \end{tabular}
    \label{tab:e3sm_strat_GOF}
\end{table}


\begin{table}[hbt!]
	\centering
 \caption{Mount Pinatubo exemplar: pathway edge weights and their standard deviations for the stratospheric warming pathway obtained using globally-averaged data windowed to 750 days.}
	\begin{tabular}{llcccc}
		\hline \hline
       	\multirow{2}{*}{\textbf{Source}} & \multirow{2}{*}{\textbf{Target}} & \textbf{Lag } & \textbf{SHAP }& \textbf{Weight }&\textbf{Ensembles }\\
   & &{\bf (days)} &{\bf Weight} & $\sigma$ & {\bf with edge} \\ 
        \hline
Globe\_T050&Globe\_T050&1&0.2500&0.0363&5\\
Globe\_FLNTC&Globe\_FLNTC&1&0.1767&0.0136&5\\
Globe\_AEROD\_v&Globe\_AEROD\_v&1&0.1347&0.0074&5\\
Globe\_T050&Globe\_FLNTC&1&0.0016&0.0003&5\\
Globe\_AEROD\_v&Globe\_FLNTC&6&0.0014&0.0007&5\\
Globe\_FLNTC&Globe\_AEROD\_v&31&0.0004&0.0003&5\\
Globe\_FLNTC&Globe\_T050&36&0.0001&0.0001&5\\
		 \hline
	\end{tabular}
	
	\label{tab:e3sm_LV_ens_edges_strat}
\end{table}

\begin{figure}[hbt!]
	\centering
	\includegraphics[width=0.45\linewidth]{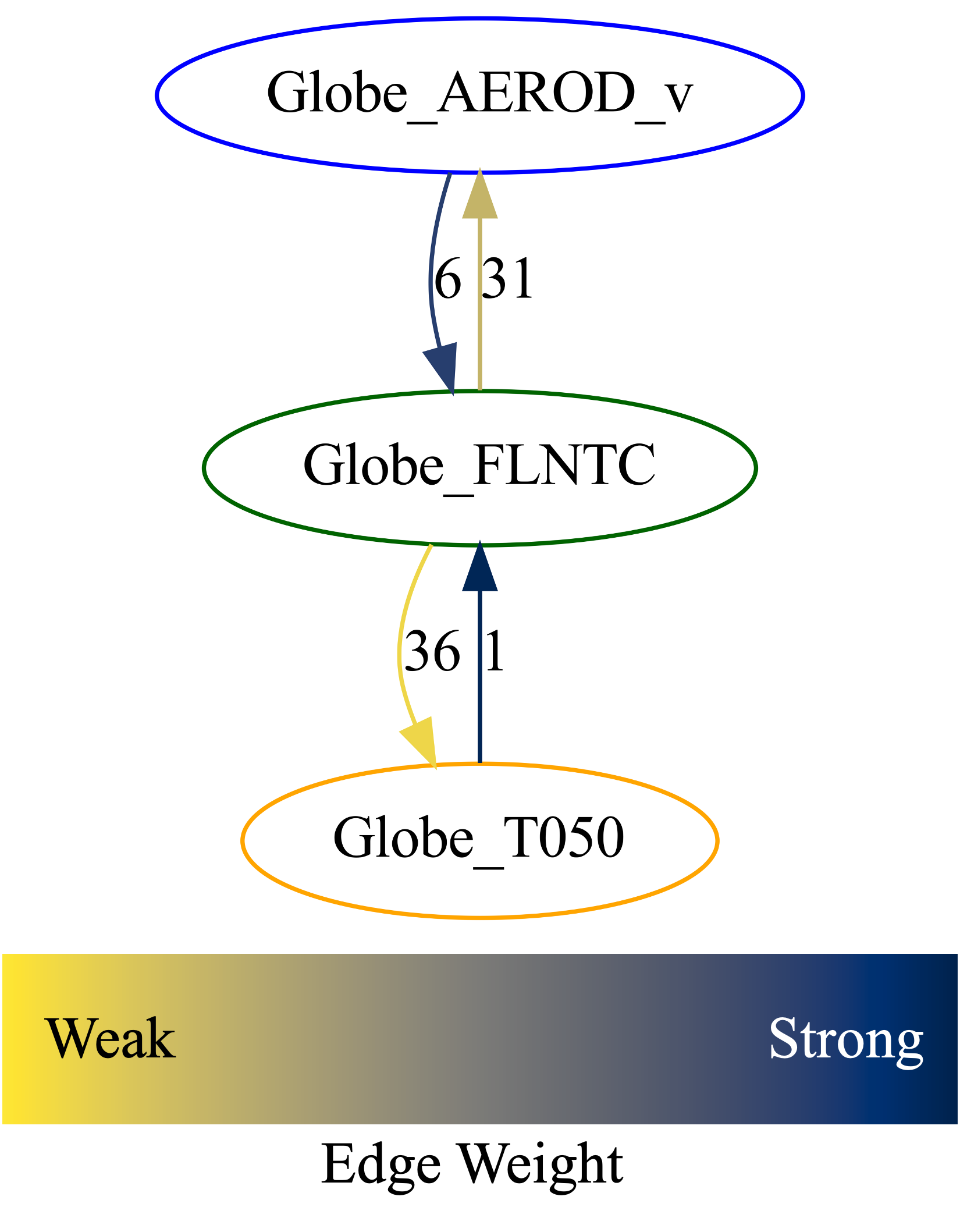}
	\caption{Mount Pinatubo exemplar: stratospheric warming pathway obtained from globally-averaged data. .
 Variable colors correspond to the colors used in the verification figure (Figure \ref{fig:e3sm_strat_GOF}). Edge weights are shown in colors ranging from blue (strong edge) to yellow (weak edge) and labeled according to the lag in days.}
	\label{fig:e3sm_LV_ens_edges_strat}
\end{figure}

Table \ref{tab:e3sm_LV_ens_edges_strat} shows the pruned pathway results for each source to target node, including the time lag, the mean and standard deviation SHAP weights, as well as the number of ensembles that contained that edge. The full results before pruning are shown in the Appendix (Table \ref{tab:app_global_strat}).
Figure \ref{fig:e3sm_LV_ens_edges_strat} shows the 
stratospheric warming pathway graph obtained using globally-averaged data and corresponding to Table \ref{tab:e3sm_LV_ens_edges_strat}.  To simplify the visualization, we exclude autocorrelated connections 
from this graph.
The variable colors correspond to the colors used in the verification figure (Figure \ref{fig:e3sm_strat_GOF}), with AEROD\_v in blue, FLNTC in green, and T050 in yellow. The edge weights are shown by the arrows connecting each node in colors ranging from blue (strong edge) to yellow (weak edge) and labeled according to the lag in days. 
Even though our pruning criteria allowed for four incoming edges to each node (Section \ref{sec:prune}), 
to simplify the visualization, we show the top three incoming edges to each node after removing any auto-correlated edges.
The reader can see a strong connection going from AEROD\_v to FLNTC with a 6-day lag and a weak connection from FLNTC to T050 with a 36-day lag.  These ``forward'' relationships are expected 
and demonstrate the 
influence of aerosols in the stratosphere on radiative forcing and temperature.  
While it took only approximately 22 days for the aerosols to encircle the globe (see Figure \ref{fig:stratospheric_burden}), the longer 36-day lag time found in the FLNTC to T050 relationship is likely due to the fact that we are working with global averages, which can be slower to respond than regional averages or individual grid cells. 
The reader can observe that there are several ``back'' edges in Figure \ref{fig:e3sm_strat_GOF}, 
that is, edges where we see the opposite of the expected relationships (e.g., T050 causing changes to AEROD\_v).  In particular, there is 
a moderate back edge from FLNTC to AEROD\_v with a 31-day lag, and a strong back edge from T050 to FLNTC with a 1-day lag.  We attribute these back edges to missing intermediary variables in our analysis.
For example, including an intermediary variable representing wind patterns could explain the T050 to FLNTC relationship: if warmer temperatures have an impact on wind patterns and therefore aerosol spread, it is conceivable for T050 to have an indirect influence on FLNTC.\\


\noindent {\bf \textit{Surface Cooling Pathway}} \\

\begin{figure}[hbt!]
    \centering
    \includegraphics[width=0.9\linewidth]{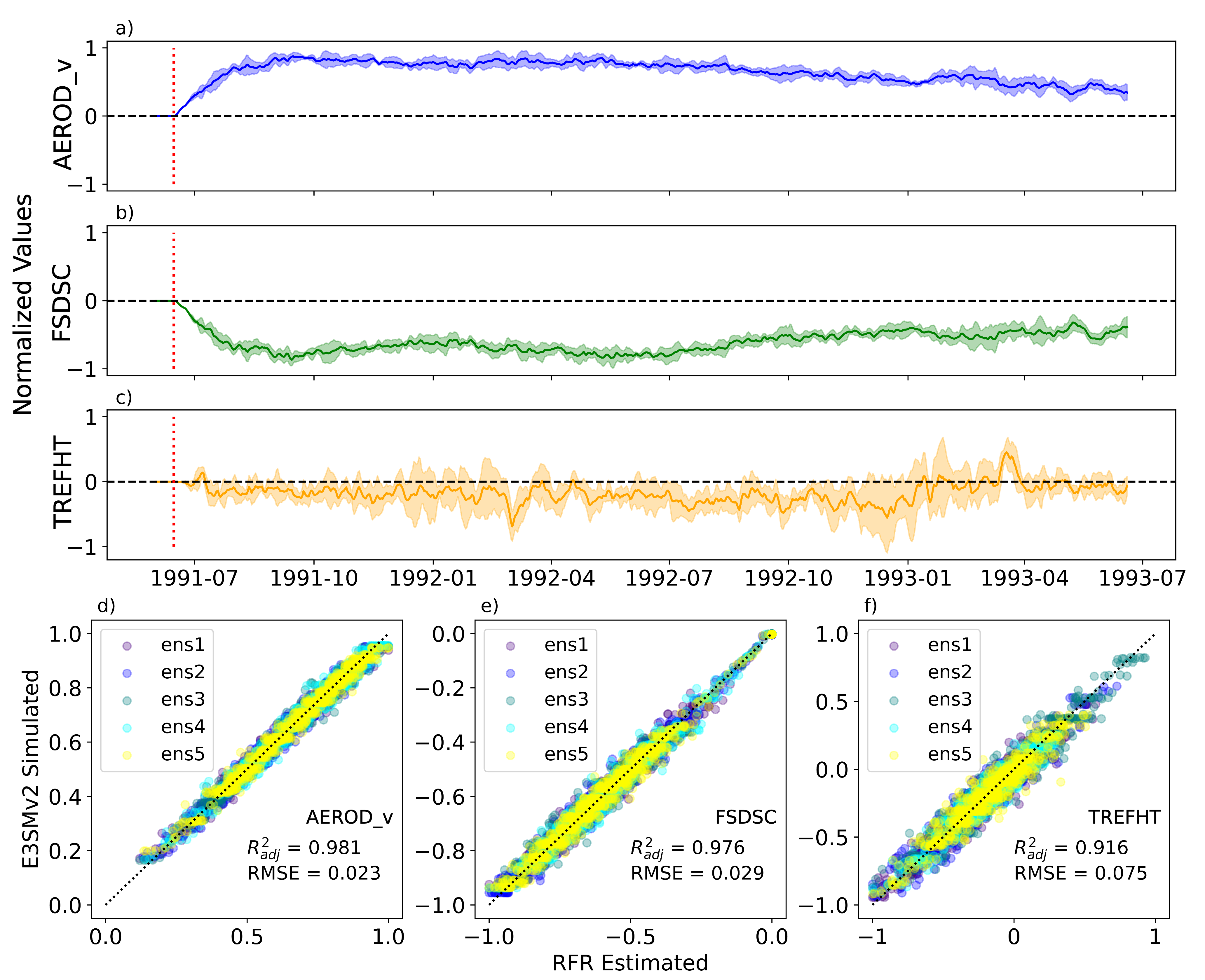}
    \caption{Mount Pinatubo exemplar: goodness of fit statistics for the surface cooling pathway using globally-averaged variables. Subplots a), b) and c) show the input time-series data simulated using E3SMv2-SPA for the variables AEROD\_v, FSDSC and TREFHT, respectively. Shaded areas show the ensemble standard deviation. The dark line is the ensemble mean.  The dashed red line shows the date of Mount Pinatubo eruption. Subplots d), e), and f) show the scatterplots between the normalized features ($y$-axis) and their RFR reconstructions ($x$-axis).
    }
    \label{fig:e3sm_surf_GOF}
\end{figure}

Next, we turn our attention to the globally averaged surface cooling pathway: AEROD\_v $\to$ FSDSC $\to$ TREFHT. 
The RFR reconstructions of the variables defining this pathway are shown in Figure \ref{fig:e3sm_surf_GOF}, and the goodness of fit statistics are given in Table \ref{tab:e3sm_surf_GOF}.  
The reader can observe that TREFHT is the noisiest variable; this makes the identification 
of the surface cooling pathway particularly difficult.  
The variable fits are slightly better than they were for the stratospheric warming pathway (Table \ref{tab:e3sm_strat_GOF}). 
We note that the $R^2_{adj}$ and $RMSE$ values for the AEROD\_v quantity are slightly different in Tables \ref{tab:e3sm_strat_GOF} and \ref{tab:e3sm_surf_GOF}; this is because our RFR fit was performed for different sets of variables in these two cases.

\begin{table}[hbt!]
    \centering
     \caption{Mount Pinatubo exemplar: goodness of fit statistics for the pruned globally-averaged data and the surface cooling pathway shown in Figure \ref{fig:e3sm_LV_ens_edges_surf}.}
    \begin{tabular}{cccccc}
    \hline \hline
         \textbf{Variable}&\textbf{$\textbf{R}^2_{adj}$ mean}& \textbf{$\textbf{R}^2_{adj}$ $\sigma$} & \textbf{$RMSE$ mean} & \textbf{$RMSE$ $\sigma$}\\
         \hline
        AEROD\_v&0.981&0.002&0.023&0.001\\
FSDSC&0.976&0.003&0.029&0.001\\
TREFHT&0.916&0.016&0.075&0.006\\
         \hline
    \end{tabular}
    \label{tab:e3sm_surf_GOF}
\end{table}

\begin{figure}[hbt!]
	\centering
	\includegraphics[width=0.45\linewidth]{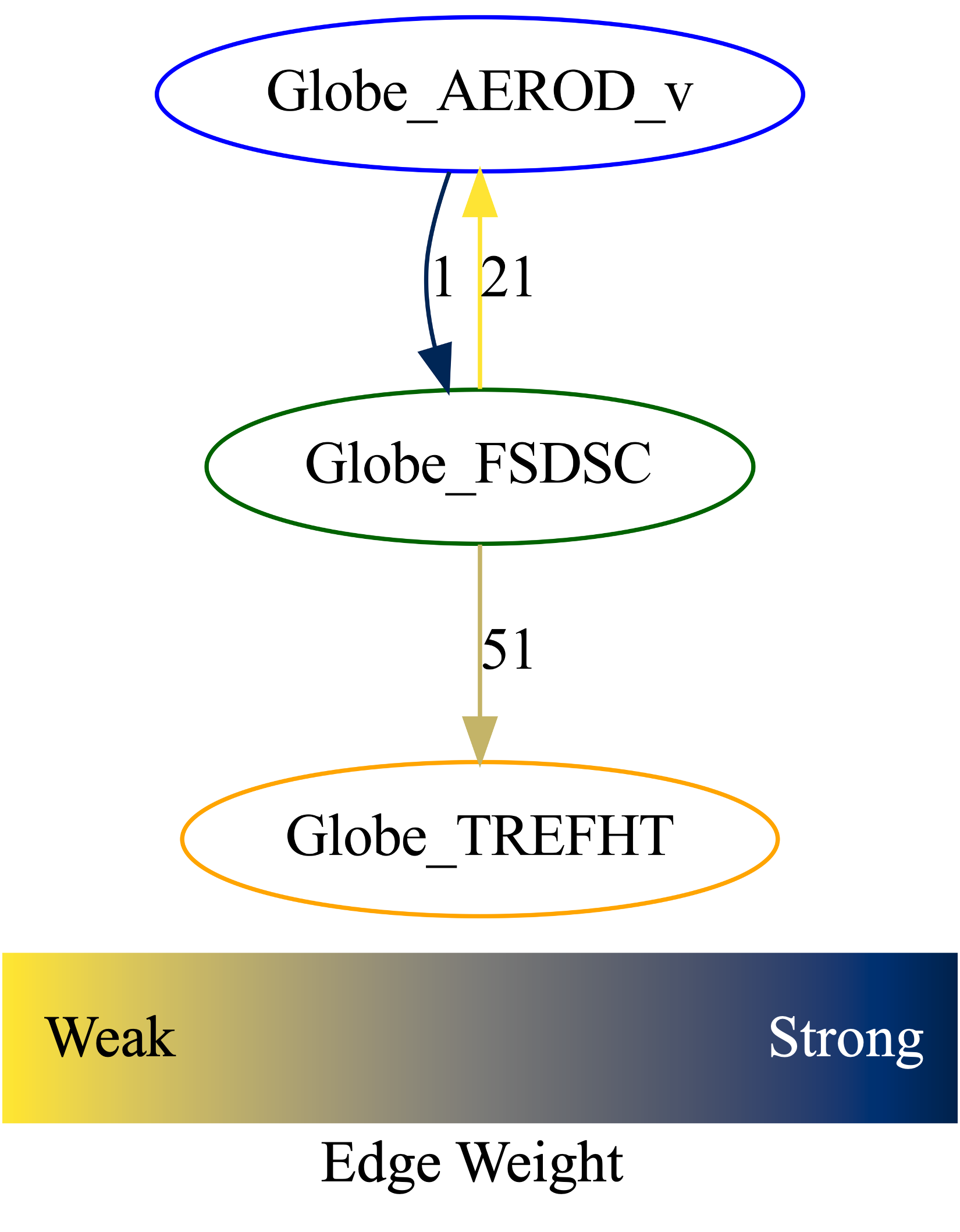}
	\caption{Mount Pinatubo exemplar: surface cooling pathway obtained from globally-averaged data. Variable colors correspond to the colors used in the verification figure (Figure \ref{fig:e3sm_surf_GOF}). Edge weights are shown in colors ranging from blue (strong edge) to yellow (weak edge) and labeled according to the lag in days.}
	\label{fig:e3sm_LV_ens_edges_surf}
\end{figure}

Figure \ref{fig:e3sm_LV_ens_edges_surf} shows the surface cooling pathway for the variables of interest, with AEROD\_v in blue, FSDSC in green, and TREFHT in yellow. As before, the edge weights are shown by the arrows, with strong edges in blue and weak edges in yellow, again excluding the autocorrelated edges. Figure \ref{fig:e3sm_LV_ens_edges_surf} shows a strong 1-day lag between AEROD\_v and FSDSC, a moderate 51-day lag between FSDSC and TREFHT and a weak 1-day back edge between FSDSC and AEROD\_v. While the former two ``forward'' relationships are expected based on what is known about the surface cooling pathway, it is difficult to corroborate the time lags associated with these variable dependencies.
As before, we believe back edges are due to processes involving variables  not explicitly included in our analysis (e.g., wind-related variables).
Table \ref{tab:e3sm_LV_ens_edges_surf} shows the ensemble mean edge strength and standard deviation between the source and target variables of the pruned surface pathway, as well as the number of ensembles that contain the edge. As with all of the pathways, the variables are strongly autocorrelated, which is expected for a time-series of Earth system variables. 

\begin{table}[hbt!]
	\centering
 \caption{Mount Pinatubo exemplar: pathway edge weights and their standard deviations for the surface cooling pathway obtained using globally-averaged data windowed to 750 days.}
	\begin{tabular}{llcccc}
	\hline \hline
		\multirow{2}{*}{\textbf{Source}} & \multirow{2}{*}{\textbf{Target}} & \textbf{Lag } & \textbf{SHAP }& \textbf{Weight }&\textbf{Ensembles }\\
   & &{\bf (days)} &{\bf Weight} & $\sigma$ & {\bf with edge} \\ 
		\hline
Globe\_TREFHT&Globe\_TREFHT&1&0.2066&0.0344&5\\
Globe\_FSDSC&Globe\_FSDSC&1&0.1479&0.0069&5\\
Globe\_AEROD\_v&Globe\_AEROD\_v&1&0.1396&0.0069&5\\
Globe\_AEROD\_v&Globe\_FSDSC&1&0.0099&0.0042&5\\
Globe\_FSDSC&Globe\_TREFHT&51&0.0022&0.0011&5\\
Globe\_FSDSC&Globe\_AEROD\_v&21&0.0001&0.0001&5\\
		 \hline
	\end{tabular}
	
	\label{tab:e3sm_LV_ens_edges_surf}
\end{table}

\subsubsection{Spatial Dimension Reduction via Averaging across Latitudinal Bands} \label{sec:e3sm_global_lat_bands}

The globally averaged pathways, while informative, are admittedly sparse, making it
difficult to interpret physically the lags between features. 
While these pathways 
show the expected influence of AEROD\_v on radiative forcing and temperature, 
the time lags associated with the pathways are not always explainable, which suggests 
that the global averaging may be masking important relationships occurring at a regional scale.
To mitigate this issue, the next step of our evaluation focuses on the two latitude bands closest to Mount Pinatubo eruption: the Tropical band (23$^{\circ}$S to 23$^{\circ}$N) and the Subtropical North band (23$^{\circ}$N to 35$^{\circ}$N). \\

\noindent {\bf \textit{Stratospheric Warming Pathway}} \\

\begin{table}[hbt!]
    \centering
     \caption{Mount Pinatubo exemplar: goodness of fit statistics for the pruned zonally-averaged data and the stratospheric warming pathway}
    \begin{tabular}{ccccc}
    \hline \hline
         \textbf{Variable}&\textbf{$\textbf{R}^2_{adj}$ mean}& \textbf{$\textbf{R}^2_{adj}$ $\sigma$} & \textbf{$RMSE$ mean} & \textbf{$RMSE$ $\sigma$}\\ 
         \hline
Tropical\_T050&0.993&0.002&0.018&0.003\\
Tropical\_AEROD\_v&0.986&0.003&0.023&0.001\\
SubtropN\_T050&0.983&0.006&0.032&0.006\\
SubtropN\_AEROD\_v&0.902&0.022&0.052&0.002\\
Tropical\_FLNTC&0.893&0.021&0.047&0.004\\
SubtropN\_FLNTC&0.829&0.044&0.098&0.008\\
         \hline
    \end{tabular}
    \label{tab:e3sm_zonal_strat_gof}
\end{table}

\begin{table}[hbt!]
	\centering
 \caption{Mount Pinatubo exemplar: pathway edge weights and their standard deviations for the stratospheric warming pathway obtained using zonally-averaged data windowed to 750 days.}
	\begin{tabular}{llcccc}
	\hline \hline
			\multirow{2}{*}{\textbf{Source}} & \multirow{2}{*}{\textbf{Target}} & \textbf{Lag } & \textbf{SHAP }& \textbf{Weight }&\textbf{Ensembles }\\
   & &{\bf (days)} &{\bf Weight} & $\sigma$ & {\bf with edge} \\ 
		\hline
Tropical\_T050&Tropical\_T050&1&0.1949&0.0488&5\\
SubtropN\_T050&SubtropN\_T050&1&0.1912&0.0439&5\\
SubtropN\_FLNTC&SubtropN\_FLNTC&1&0.1892&0.0255&5\\
Tropical\_AEROD\_v&Tropical\_AEROD\_v&1&0.1562&0.0195&5\\
SubtropN\_AEROD\_v&SubtropN\_AEROD\_v&1&0.1404&0.0213&5\\
Tropical\_FLNTC&Tropical\_FLNTC&1&0.1161&0.0108&5\\
SubtropN\_AEROD\_v&SubtropN\_FLNTC&1&0.0038&0.0013&5\\
Tropical\_FLNTC&SubtropN\_FLNTC&1&0.0033&0.0013&5\\
Tropical\_AEROD\_v&Tropical\_FLNTC&6&0.0025&0.0005&5\\
Tropical\_AEROD\_v&SubtropN\_AEROD\_v&6&0.0025&0.0009&5\\
Tropical\_AEROD\_v&SubtropN\_FLNTC&46&0.0017&0.0005&5\\
Tropical\_FLNTC&SubtropN\_AEROD\_v&1&0.0013&0.0008&5\\
SubtropN\_T050&SubtropN\_AEROD\_v&46&0.0011&0.0004&5\\
SubtropN\_AEROD\_v&Tropical\_FLNTC&6&0.0009&0.0005&5\\
SubtropN\_FLNTC&Tropical\_FLNTC&1&0.0009&0.0004&5\\
Tropical\_T050&SubtropN\_T050&21&0.0006&0.0004&5\\
Tropical\_AEROD\_v&SubtropN\_T050&31&0.0004&0.0002&5\\
SubtropN\_AEROD\_v&Tropical\_AEROD\_v&21&0.0001&0.0001&5\\
		 \hline
	\end{tabular}
	
	\label{tab:e3sm_LV_ens_edges_strat_trop}
\end{table}

\begin{figure}[hbt!]
    \centering
    \includegraphics[width=0.9\linewidth]{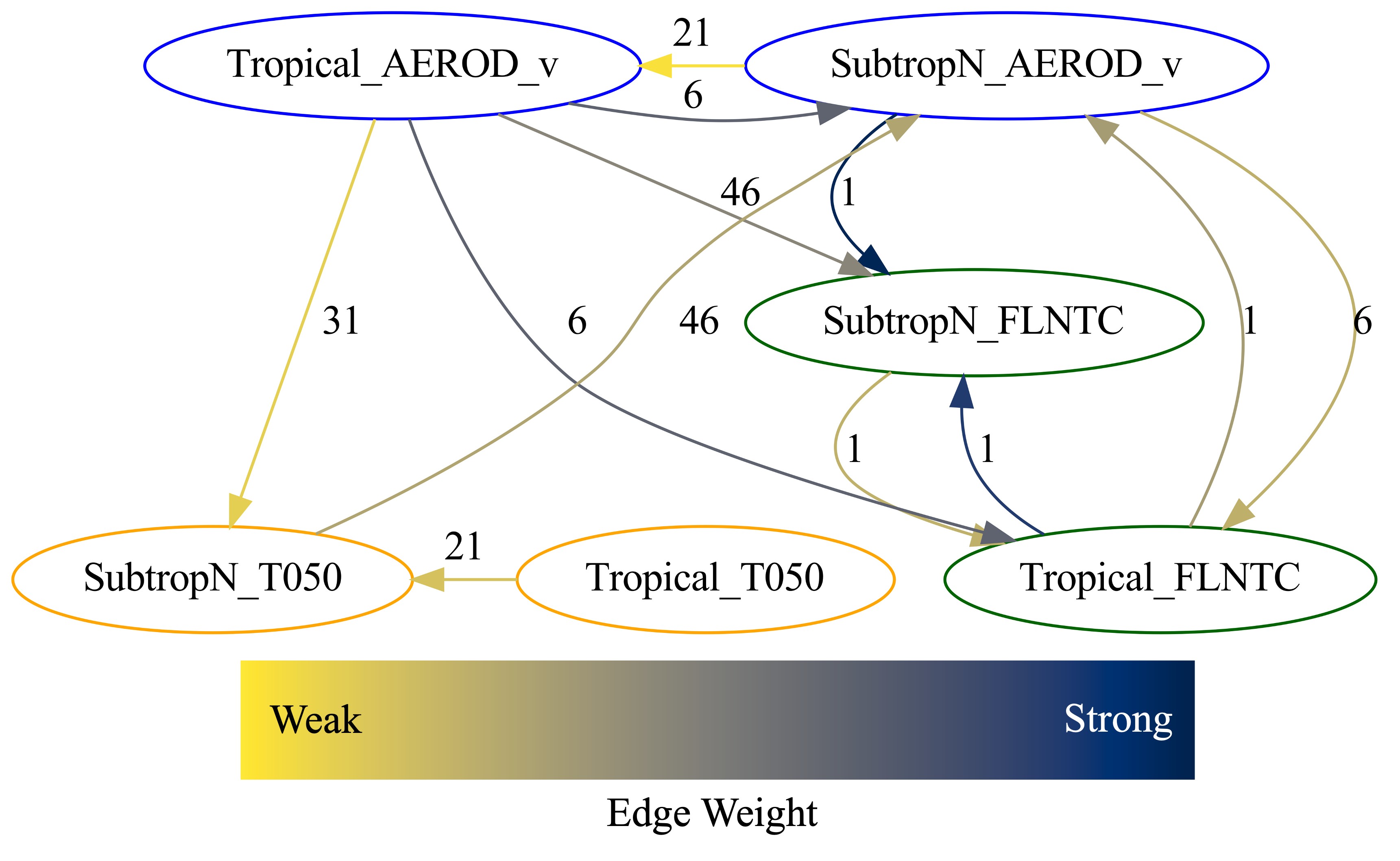}
    \caption{Mount Pinatubo exemplar: stratospheric warming pathway obtained from zonally-averaged data.. Variable colors correspond to the colors used in the verification figure (Figure \ref{fig:e3sm_strat_GOF}). Edge weights are shown in colors ranging from blue (strong edge) to yellow (weak edge) and labeled according to the lag in days.}
    \label{fig:p1_t50_tropics_graph_strat}
\end{figure}

Table \ref{tab:e3sm_zonal_strat_gof} shows the reconstruction statistics for the latitudinal bands in the stratospheric warming pathway. Generally, the reconstructions are better for the features in the Tropical band than the Subtropical North band.  As expected, the noisy FLNTC signal is the most difficult to reconstruct with high accuracy.
Table \ref{tab:e3sm_LV_ens_edges_strat_trop} shows the pruned pathway results for the two latitudinal bands.  
Figure \ref{fig:p1_t50_tropics_graph_strat} shows the pathway graph inferred from Table \ref{tab:e3sm_LV_ens_edges_strat_trop}, with autocorrelations removed for visualization purposes, 
as before. 
The variable colors correspond to the colors used in the verification 
figure (Figure \ref{fig:e3sm_LV_ens_edges_strat}).
The reader can observe that, while 
Figure \ref{fig:p1_t50_tropics_graph_strat} is much more complex than our previous analogous figure
obtained using global averages (Figure \ref{fig:e3sm_LV_ens_edges_strat}), 
the same characteristic connections between AEROD\_v, radiative flux and temperature exist. 
The strongest connections are 
those that go into the Subtropical North FLNTC from AEROD\_v in both latitudinal bands and from Tropical FLNTC.
In addition to intra-zonal band relationships (e.g., AEROD\_v$\to$ FLNTC), 
we also see some cross-band connections, many of which show a dependence of a Subtropical band variable on a Tropical band variable.  These relationships are expected, 
as they demonstrates a general northward flowing direction from the Tropics to the Subtropical North, which follows the evolution of the volcanic plume 
that was observed
after the Mount Pinatubo eruption (see Figure \ref{fig:stratospheric_burden}).
It is interesting to remark that, with our current set of pruning criteria, there are no connections from FLNTC to T050, and only a weak connection having a 31-day time lag from Tropical AEROD\_v to Subtropical North T050.  This result is surprising, given the prevalence of the FLNTC to T050 relationship 
in our globally-averaged analyses 
(Figure \ref{fig:e3sm_LV_ens_edges_strat}). \\ 

\noindent {\bf \textit{Surface Cooling Pathway}} \\

\begin{table}[hbt!]
    \centering
     \caption{Mount Pinatubo exemplar: goodness of fit statistics for the pruned zonally-averaged data and the surface cooling pathway}
    \begin{tabular}{ccccc}
    \hline \hline
         \textbf{Variable}&\textbf{$\textbf{R}^2_{adj}$ mean}& \textbf{$\textbf{R}^2_{adj}$ $\sigma$} & \textbf{$RMSE$ mean} & \textbf{$RMSE$ $\sigma$}\\ 
         \hline
Tropical\_AEROD\_v&0.986&0.003&0.023&0.001\\
Tropical\_FSDSC&0.978&0.005&0.025&0.003\\
SubtropN\_AEROD\_v&0.915&0.017&0.051&0.002\\
Tropical\_TREFHT&0.909&0.038&0.023&0.001\\
SubtropN\_FSDSC&0.845&0.042&0.069&0.009\\
SubtropN\_TREFHT&0.771&0.036&0.105&0.010\\
         \hline
    \end{tabular}
    \label{tab:e3sm_zonal_surf_gof}
\end{table}

\begin{figure}[hbt!]
    \centering
    \includegraphics[width=0.9\linewidth]{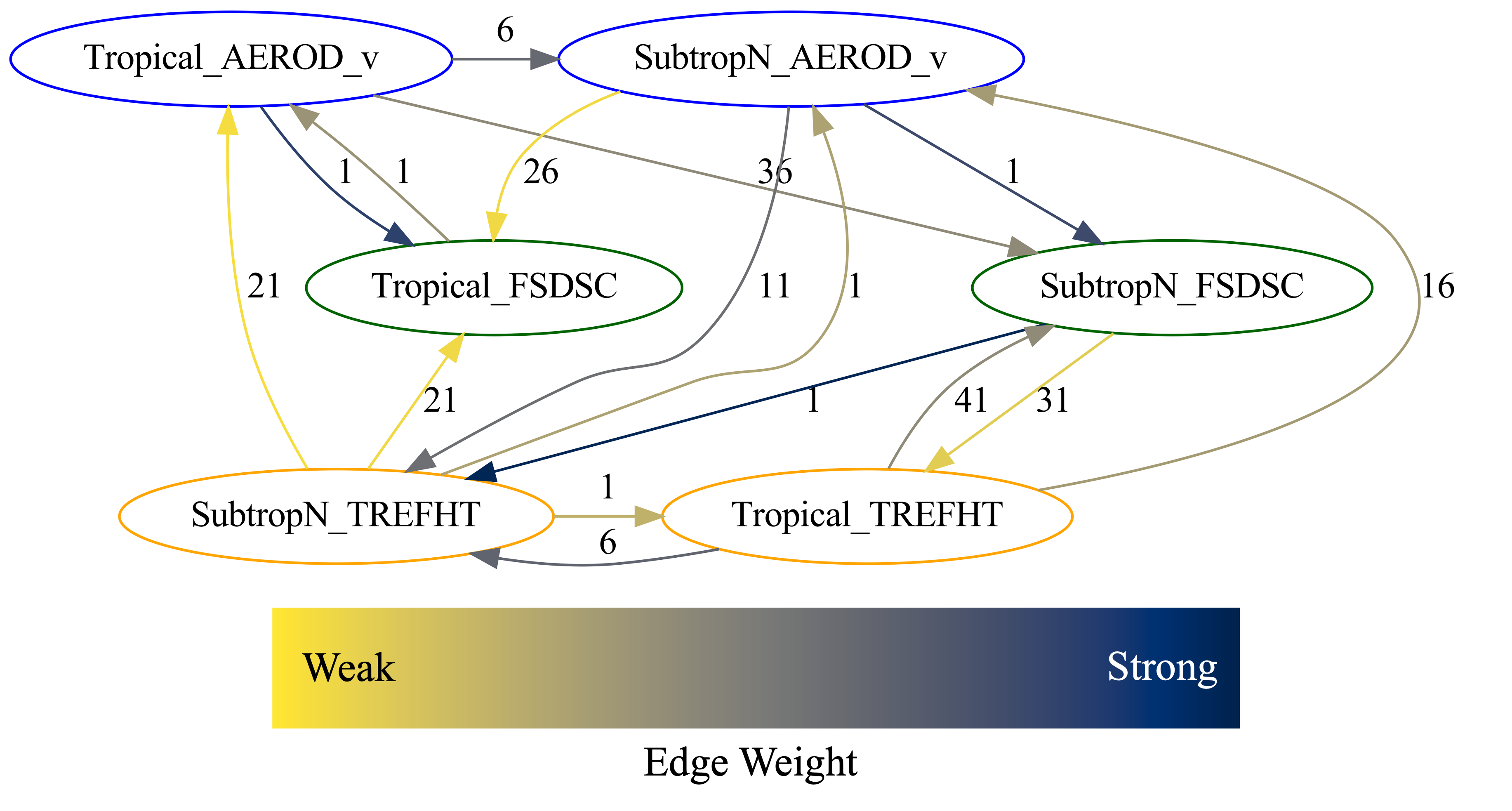}
    \caption{Mount Pinatubo exemplar: surface cooling pathway obtained from zonally-averaged data..  Variable colors correspond to the colors used in the verification figure (Figure \ref{fig:e3sm_surf_GOF}). Edge weights are shown in colors ranging from blue (strong edge) to yellow (weak edge) and labeled according to the lag in days.}
    \label{fig:p1_trefht_tropics_graph_surf}
\end{figure}

When we look at the surface cooling pathway for the latitudinally averaged features, we expect to see a similar northward influence, and the connections between AEROD\_v $\to$ FSDSC $\to$ TREFHT that we saw in the globally averaged pathway. 
Table \ref{tab:e3sm_zonal_surf_gof} shows the reconstruction statistics for the surface cooling pathway.
As in the stratospheric warming pathway, the Subtropical North variables are more difficult to reconstruct than the Tropical variables, with the Subtropical North TREFHT variable having the lowest $R^2_{adj}$.
The pruned pathway can be inferred from Table \ref{tab:e3sm_LV_ens_edges_surf_trop} 
and is plotted in Figure \ref{fig:p1_trefht_tropics_graph_surf} with autocorrelations removed. 
Again, we see a more complex relationship between the variables than in the global surface cooling pathway graph (Figure \ref{fig:e3sm_LV_ens_edges_surf}).  The reader can observe
the presence of the expected
characteristic patterns of connections going from AEROD\_v to FSDSC and to TREFHT, as well as the expected northward propagation of the aerosols and their impacts from the Tropical band to the Subtropical North band. 
The time lags associated with many of the relationships uncovered are on the order of 21-36 days, which is consistent with the time it took for the aerosols from Mount Pinatubo to encircle the globe (Figure \ref{fig:stratospheric_burden}).  

\begin{table}[hbt!]
	\centering
 \caption{Mount Pinatubo exemplar: pathway edge weights and their standard deviations for the surface cooling pathway obtained using zonally-averaged data windowed to 750 days.}
	
	\begin{tabular}{llcccc}
	\hline \hline
			\multirow{2}{*}{\textbf{Source}} & \multirow{2}{*}{\textbf{Target}} & \textbf{Lag } & \textbf{SHAP }& \textbf{Weight }&\textbf{Ensembles }\\
   & &{\bf (days)} &{\bf Weight} & $\sigma$ & {\bf with edge} \\ 
		\hline
SubtropN\_TREFHT&SubtropN\_TREFHT&1&0.1735&0.0099&5\\
Tropical\_AEROD\_v&Tropical\_AEROD\_v&1&0.1559&0.0197&5\\
SubtropN\_AEROD\_v&SubtropN\_AEROD\_v&1&0.1474&0.0200&5\\
SubtropN\_FSDSC&SubtropN\_FSDSC&1&0.1435&0.0198&5\\
Tropical\_FSDSC&Tropical\_FSDSC&1&0.1368&0.0319&5\\
Tropical\_TREFHT&Tropical\_TREFHT&1&0.0682&0.0181&5\\
SubtropN\_FSDSC&SubtropN\_TREFHT&1&0.0045&0.0010&5\\
Tropical\_AEROD\_v&Tropical\_FSDSC&1&0.0038&0.0020&5\\
SubtropN\_AEROD\_v&SubtropN\_FSDSC&1&0.0036&0.0012&5\\
Tropical\_TREFHT&SubtropN\_TREFHT&6&0.0029&0.0015&5\\
Tropical\_AEROD\_v&SubtropN\_AEROD\_v&6&0.0028&0.0018&5\\
SubtropN\_AEROD\_v&SubtropN\_TREFHT&11&0.0026&0.0006&5\\
Tropical\_AEROD\_v&SubtropN\_FSDSC&36&0.0020&0.0007&5\\
Tropical\_TREFHT&SubtropN\_FSDSC&41&0.0019&0.0011&5\\
Tropical\_FSDSC&Tropical\_AEROD\_v&1&0.0017&0.0015&5\\
Tropical\_TREFHT&SubtropN\_AEROD\_v&16&0.0016&0.0010&5\\
SubtropN\_TREFHT&SubtropN\_AEROD\_v&1&0.0014&0.0007&5\\
SubtropN\_TREFHT&Tropical\_TREFHT&1&0.0011&0.0008&5\\
SubtropN\_FSDSC&Tropical\_TREFHT&31&0.0005&0.0001&5\\
SubtropN\_TREFHT&Tropical\_FSDSC&21&0.0003&0.0002&5\\
SubtropN\_AEROD\_v&Tropical\_FSDSC&26&0.0003&0.0002&5\\
SubtropN\_TREFHT&Tropical\_AEROD\_v&21&0.0002&0.0001&5\\
		 \hline
	\end{tabular}
	\label{tab:e3sm_LV_ens_edges_surf_trop}
\end{table}

\section{Conclusions and Future Work}\label{sec:conclusion}

The primary contribution of this paper is the development of a data-driven ``outer loop'' algorithm for teasing out source-impact relationships in large climate datasets.
Whereas traditional ML approaches employ RFR and feature importance for classification or regression, 
our unique combination of these two methods enables us to extract from climate data the relative 
influence of one feature on another, towards tracing out source-impact pathways.
 After developing some quantitative methods for evaluating our approach and 
 verifying it on a set of synthetic coupled equations, we deployed the method on two known pathways that occurred as a result of the 1991 eruption of Mount Pinatubo: the stratospheric warming and surface cooling pathways.  
In order to make the method tractable for this complex high-dimensional exemplar, data reduction 
in the form of spatial averaging and temporal binning was performed.  

While our approach was generally successful in finding the key ``nodes'' in each of the sought-after
temperature pathways, it also identified some unexpected relationships.  
Our analysis revealed that reducing the data via a global averaging  can over-smooth the data, making it more difficult for our method to pick up on the signals of interest, and complicating the expected temporal relationships between features. For both the global and zonal averaging techniques considered in this paper, we observed the presence of ``back'' edges in our pathway graphs, i.e., in addition to determining that FLNTC is influenced by AEROD\_v, as expected, our analysis found that AEROD\_v is also influenced by 
FLNTC. 
We believe that, when this happens, our method is finding connections involving  variables and  physical mechanisms 
not utilized explicitly in the pathway analysis performed, e.g., wind field information.  Repeating our analysis with additional input variables to confirm this conjecture would be an interesting future 
research endeavor. 
We additionally intend to do more regional analysis, e.g., by repeating the analysis performed herein using IPCC regions \cite{IPCC:2020}. 
Finally, for the zonal analysis, we found that our method did not always 
discover the expected time-lags of influence between the relevant variables/features. 
To try to improve on this, future work will involve more substantial windowing of the data, 
in particular, using sliding windows to study the temporal system-wide dynamic impacts of a climate source.  

While we have demonstrated the method in a confirmatory capacity by postulating a set of relationships and using the method to confirm/deny them, our method has the potential to \textit{discover} source-impact pathways directly from data. Deploying the method in this \textit{exploratory} capacity may enable the discovery of previously unknown relationships in the climate.  Finally, it is important to recognize that the relationships identified by our approach are correlative rather than causal.  Developing a causal analysis-based method that can pick out  causal vs. correlative edges in our pathway graphs would be of tremendous interest.

\section*{Acknowledgments}

This material is based upon work supported by the Laboratory Directed Research and Development (LDRD) program at Sandia National Laboratories. 
The research used resources of the National Energy Research Scientific Computing Center (NERSC), a Department of Energy Office of Science User Facility using NERSC award BER-ERCAP0026535.  The writing of this manuscript was funded in part by the third author’s (Irina Tezaur's) Presidential Early Career Award for Scientists and Engineers (PECASE).

This article has been authored by an employee of National Technology \& Engineering Solutions of Sandia, LLC under Contract No. DE-NA0003525 with the U.S. Department of Energy (DOE). The employee owns all right, title and interest in and to the article and is solely responsible for its contents. The United States Government retains and the publisher, by accepting the article for publication, acknowledges that the United States Government retains a non-exclusive, paid-up, irrevocable, world-wide license to publish or reproduce the published form of this article or allow others to do so, for United States Government purposes. The DOE will provide public access to these results of federally sponsored research in accordance with the DOE Public Access Plan \url{https://www.energy.gov/downloads/doe-public-access-plan}.

The authors would like to thank Hunter Brown, Benjamin Wagman, Thomas Ehrmann, and Joe Hollowed for providing invaluable subject matter expertise in stratospheric dynamics and climate modeling, which enabled 
the interpretations discussed within this paper.  We would also like to thank Laura Swiler for helping us refine various details in our algorithm, including the choice of pruning criteria.

\section*{Data and code availability}

Data from the full E3SMv2-SPA simulation campaign including pre-industrial control, historical, and Mount Pinatubo ensembles will be hosted at Sandia National Laboratories with location and download instructions announced on \url{https://www.sandia.gov/cldera/e3sm-simulations-data/} when available.  
A Python code implementing our RFR and feature importance-based approach is currently awaiting copyright assertion, and will be made publicly available as soon as this is possible.
The E3SMv2-SPA climate code that was used to generate the data analyzed herein can be found at \url{https://github.com/sandialabs/CLDERA-E3SM}. 

\bibliographystyle{elsarticle-num}
\bibliography{references}

\end{document}